\documentclass[10pt,twocolumn,letterpaper]{article}

\usepackage[pagenumbers]{cvpr} 

\usepackage[dvipsnames]{xcolor}
\usepackage{amsmath}
\usepackage{amsthm}
\usepackage{amsfonts}
\usepackage{amssymb}
\usepackage{graphicx}
\usepackage{adjustbox}
\usepackage{multicol}
\usepackage{mathtools}
\usepackage{enumitem}

\usepackage{bm}
\usepackage{caption}
\usepackage{subcaption}
\usepackage{multirow}
\usepackage{tikz}

\newcommand{\R}{\mathbb{R}}

\newcommand{\name}{DART}
\definecolor{originblue}{RGB}{68,114,196}

\definecolor{cvprblue}{rgb}{0.21,0.49,0.74}
\usepackage[pagebackref,breaklinks,colorlinks,citecolor=cvprblue]{hyperref}

\def\paperID{13119} 
\def\confName{CVPR}
\def\confYear{2024}

\title{DART: Implicit Doppler Tomography for Radar Novel View Synthesis}

\author{
Tianshu Huang\textsuperscript{*1}
\hspace{1.5em} John Miller\textsuperscript{*12}
\hspace{1.5em} Akarsh Prabhakara\textsuperscript{1}
\hspace{1.5em} Tao Jin\textsuperscript{1}
\hspace{1.5em} Tarana Laroia\textsuperscript{1}
\\
Zico Kolter\textsuperscript{12}
\hspace{1.5em} Anthony Rowe\textsuperscript{12}
\\
\textsuperscript{1}Carnegie Mellon University
\hspace{1.5em}\textsuperscript{2}Bosch Research
\\
{\tt\small \{tianshu2, jmiller4, aprabhak, taojin, tlaroia, agr, zkolter\}@andrew.cmu.edu}
}

\begin{document}
\maketitle

\begingroup\renewcommand\thefootnote{*}
\footnotetext{Equal Contribution.}
\endgroup

\begin{abstract}
Simulation is an invaluable tool for radio-frequency system designers that enables rapid prototyping of various algorithms for imaging, target detection, classification, and tracking. However, simulating realistic radar scans is a challenging task that requires an accurate model of the scene, radio frequency material properties, and a corresponding radar synthesis function. Rather than specifying these models explicitly, we propose DART --- Doppler Aided Radar Tomography, a Neural Radiance Field-inspired method which uses radar-specific physics to create a reflectance and transmittance-based rendering pipeline for range-Doppler images. We then evaluate DART by constructing a custom data collection platform and collecting a novel radar dataset together with accurate position and instantaneous velocity measurements from lidar-based localization. In comparison to state-of-the-art baselines, DART synthesizes superior radar range-Doppler images from novel views across all datasets and additionally can be used to generate high quality tomographic images.\footnote{
Our implementation, data collection platform, and collected datasets can be found via our project site: \url{https://wiselabcmu.github.io/dart/}.
}
\end{abstract}

\section{Introduction}

Driven by advances in the automotive industry, miniaturized millimeter wave (mmWave) radar chips are becoming cheaper and more ubiquitous. Boasting a high range resolution and the ability to penetrate light materials, mmWave radars have proven effective in many application domains including collision avoidance and driver assistance in automobiles \cite{russell1997millimeter,Qian_2021_CVPR,Li_2022_CVPR_1,Scheiner_2020_CVPR,Ding_2023_CVPR}, through-occlusion imaging in airport scanners \cite{sheen1998cylindrical,8102042}, and vision-denied tracking and mapping \cite{cen2018precise, lu2020milliego, almalioglu2020milli,Li_2022_CVPR_2,Guan_2020_CVPR}.

Because designing, testing, and deploying new radar systems in the real world can be costly, many rapid prototyping pipelines heavily rely on simulation. Modern radar simulation tools normally require the user to manually specify the geometry and characteristics of the scene, including all material properties \cite{auer2016raysar}. While other sensors (e.g. lidar) can be used to scan an environment and produce a mesh or voxel map, they cannot capture radar-specific material properties that are crucial for generating realistic radar scans. Thus, in practice, this results in greatly simplified environment models due to the difficulty of meticulously surveying a scene and generating (or annotating) a model by hand. 

\begin{figure}
    \includegraphics[width=\columnwidth]{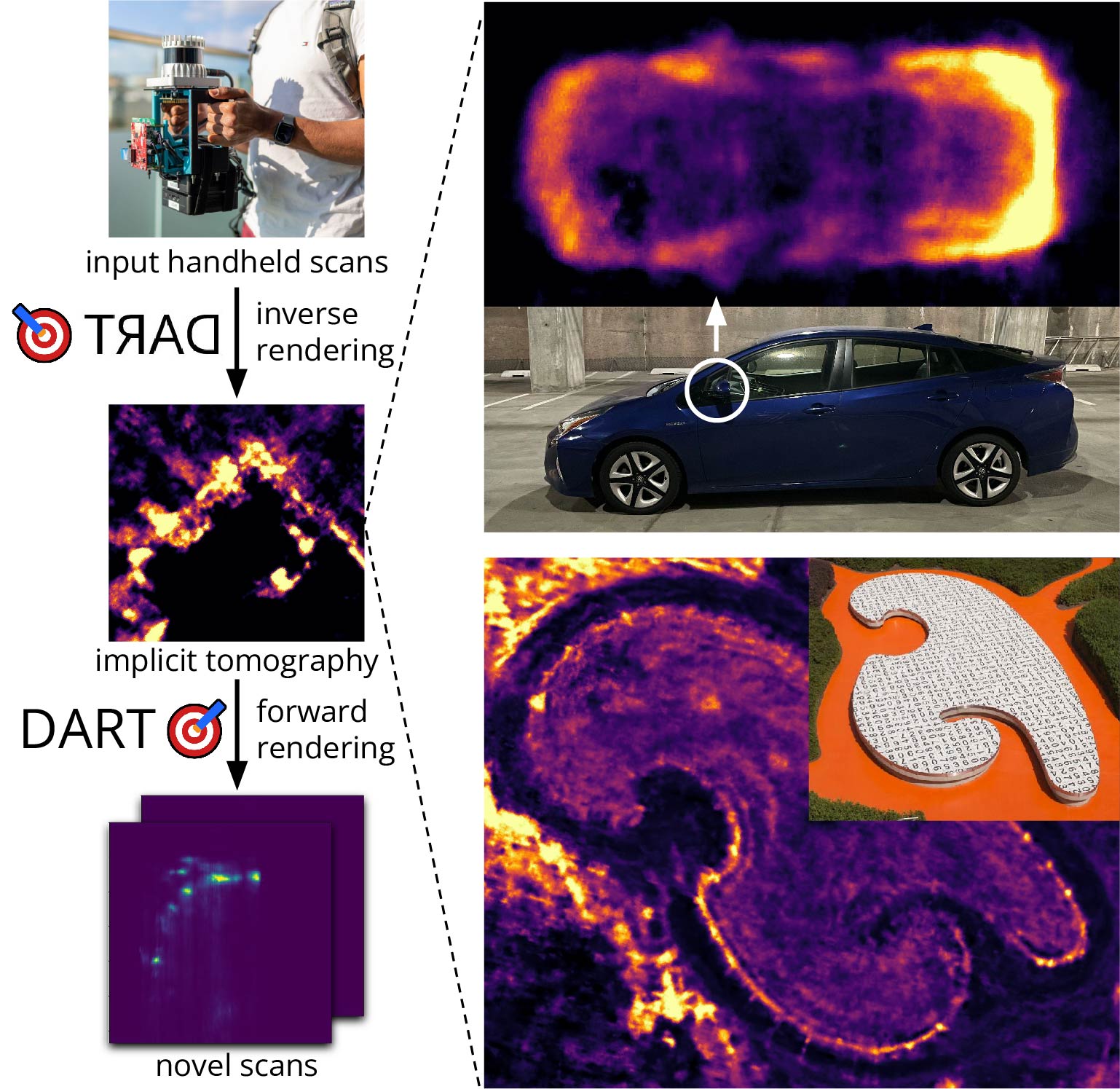}
    \vspace*{-1.5em}
    \caption{DART uses scans from a handheld radar to learn an implicit tomography of a scene in order to accurately render scans from novel viewpoints (left). DART's implicit tomography can also be sampled to map the radar properties of a scene (right).}
    \vspace{-1.0em}
    \label{fig:hero}
\end{figure}

We envision a more intelligent, data-driven approach to scene modeling for radar simulation where a user can carry a handheld radar sensor through a static environment and automatically generate a model suitable for accurate simulation of that environment. To this end, we frame the radar simulation problem as one of \emph{novel view synthesis}: using several radar measurements of a scene to simulate what a radar would see from a new pose. Such a system would not only accelerate the development and testing of new algorithms across a variety of environmental conditions, but also open the door to a myriad of new inference techniques in radar sensing such as localization, mapping, imaging, and recognition which rely on accurate forward rendering models and could greatly benefit from realistic radar models.

\paragraph{Novel View Synthesis} Neural Radiance Fields (NeRFs) \cite{mildenhall2021nerf} have revolutionized novel view synthesis, leading to an explosion in interest in graphics and beyond. By leveraging a (neural) \emph{implicit} scene representation instead of explicitly modeling scene geometry, textures, and materials, NeRFs are able to capture and reproduce visual intricacies such as specularity, translucency, reflections, and complex occlusions. This results in a 3D scene capture and rendering system that boasts an unprecedented level of photorealism.


Drawing inspiration from the success of NeRFs, we formulate an analogous problem for mmWave radar imaging. Our method, \textit{Doppler-Aided Radar Tomography} (DART), takes a similar approach by implicitly capturing material properties from input scans which are reproduced when the model is sampled from a novel viewpoint. Though our model is implicit, we can also generate an explicit tomographic image by sampling along a voxel grid, which we use to show that \name\ is not simply memorizing the input data, but is in fact \emph{learning} the geometry and material properties of the scene (Fig.~\ref{fig:hero}).

\paragraph{Key Challenges} Applying NeRF's implicit scene modeling paradigm to the radar domain presents substantial challenges. We derive a rendering model from the ground up that appropriately reflects the unique nature of radar wave propagation. In NeRF, rendering each pixel involves integrating samples along a 1D ray, following a pinhole camera model \cite{mildenhall2021nerf}. However, radar waves propagate \emph{radially} from the antenna. Even after range-azimuth-elevation processing, each radar pixel corresponds to a coarse 2D region of space, as the elevation-azimuth resolution of compact mmWave radars tends to be relatively poor\footnote{For context, these radars have angular resolutions on the order of 15$^{\circ}$, orders of magnitude worse than cameras ($\approx$0.01$^{\circ}$) \cite{Rebut_2022_CVPR,ti:AWR1843AOP}}. One key insight is to choose a radar representation space --- range-Doppler --- which greatly reduces angular ambiguity in one dimension under the assumption that the scene is static and the radar is moving with a known velocity \cite{hlawatsch1998time}. This presents additional systems challenges, as the sensor platform needs to be moving and its velocity must be measured accurately alongside its position and orientation.

\begin{figure}
\centering
\begin{subfigure}[c]{.33\columnwidth}
    \centering
    \includegraphics[width=\columnwidth]{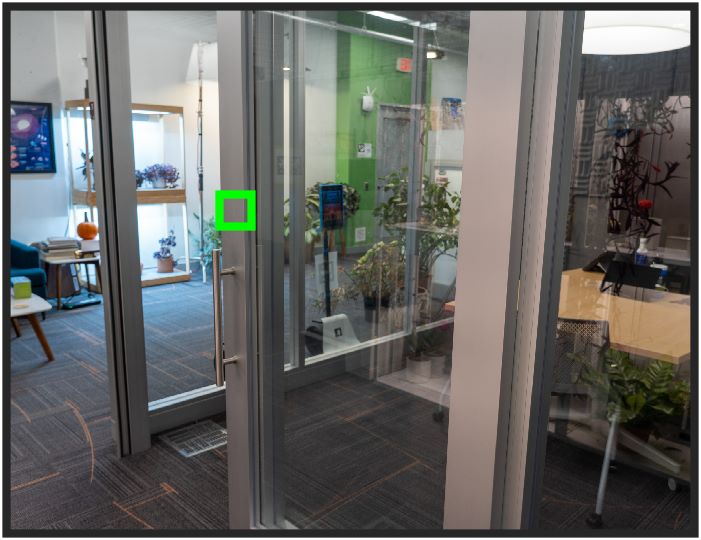}
\end{subfigure}
\begin{subfigure}[c]{.33\columnwidth}
    \centering
    \includegraphics[width=\columnwidth]{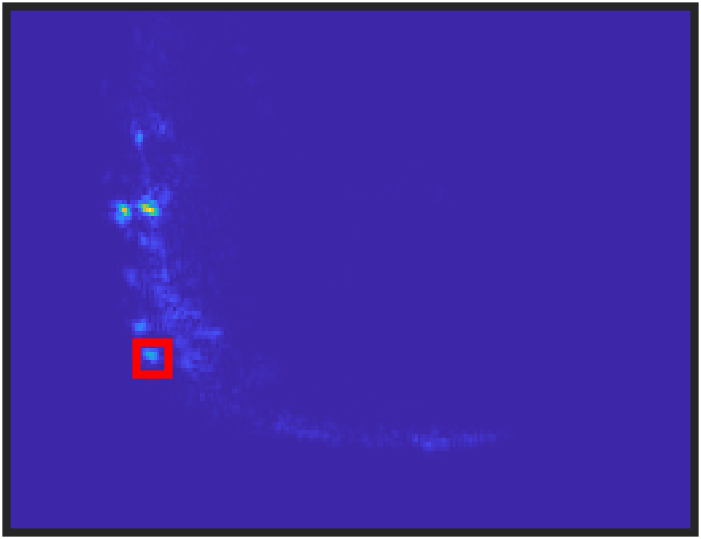}
\end{subfigure}
\begin{subfigure}[c]{.27\columnwidth}
    \centering
    \includegraphics[width=\columnwidth]{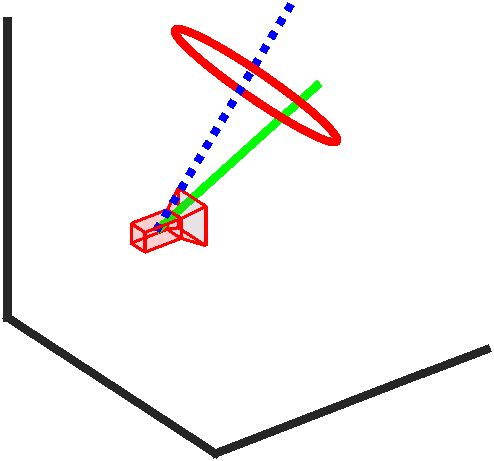}
\end{subfigure}
\vspace{-0.5em}
\caption{NeRF's pinhole camera model renders a pixel (left) by integrating along a ray (right, green), while DART's range-Doppler model renders a pixel (middle) by integrating along a velocity-dependent (right, blue) circle (right, red).}
\vspace{-0.5em}
\label{fig:toy_rangedoppler}
\end{figure}

Even with the dimensionality reduction afforded by range-Doppler processing, rendering a single radar pixel involves integrating samples along a \emph{circle}, rather than a ray (Fig. \ref{fig:toy_rangedoppler}). However, appropriately capturing occlusion effects requires that the nearest ranges are processed first due to occlusion caused by objects closer to the radar. Additionally, the size of the integration arc grows as the distance from the radar increases, resulting in an effective decrease in sampling density for points further from the radar that needs to be accounted for. Through careful modeling of these effects and a clever sampling scheme prioritizing sample re-use, we derive a computationally efficient forward rendering function that produces realistic novel radar scans.

\paragraph{Contributions} We propose DART: Doppler-Aided Radar Tomography, which implicitly learns a tomographic representation of the world in order to accurately synthesize radar range-Doppler images. To summarize our contributions:
\begin{enumerate}[noitemsep,topsep=0pt,leftmargin=*]
    \item We formulate the problem of radar novel view synthesis from implicit reflectance and transmittance maps using range-Doppler images.
    \item Using a NeRF-inspired technique, we explicitly formulate the forward rendering of range-Doppler radar images and implicitly invert it using gradient descent to learn a neural-implicit representation.
    \item We construct a data collection rig and collect novel radar imaging datasets with accurate position and instantaneous velocity along with reference lidar point clouds.
    \item We evaluate DART across a range of scenarios and show that it out-performs the state-of-the-art, quantitatively and qualitatively, in both its synthetic radar renderings and its implicit imaging of scenes.
\end{enumerate}

\paragraph{Limitations} Since we rely on Doppler, our method is limited to static scenes, and requires accurate velocity estimates and a constantly moving radar. While motion is intrinsic to our method, we believe that it is reasonable to require movement during scanning. Poor velocity estimates or non-static scenes can cause \name\ to perform poorly; we hope to relax these limitations in the future.

\section{Related Work}

\begin{figure*}
    \centering
    \includegraphics[width=\textwidth]{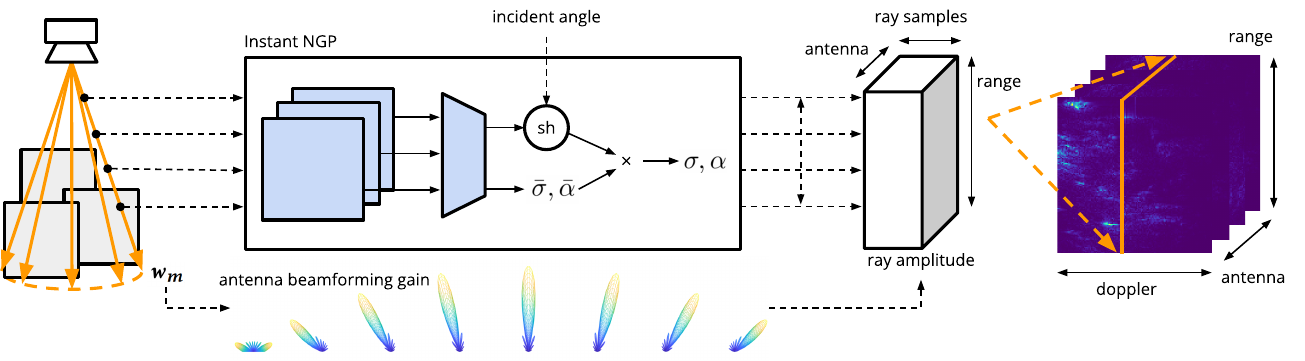}
    \vspace{-2em}
    \caption{\name\ tackles radar novel view synthesis by learning a neural implicit map of the world from a trajectory of radar measurements. We make key radar-specific decisions in choosing (1) a high quality radar representation space --- \textbf{Range-Doppler}, (2) a world model that captures radar interactions --- \textbf{$\sigma$ and $\alpha$ with spherical harmonics coefficients}, (3) a network architecture to model our desired representation --- \textbf{Instant NGP}, and (4) an optimized radar rendering and training methodology --- \textbf{Range-Doppler specific rendering}.}
    \label{fig:dart-overview}
    \vspace{-0.5em}
\end{figure*}

\subsection{Radar Simulation}
\label{sec:radar-simulation}

\paragraph{Model-Based Approaches} \textit{Model-based} methods use a physics and environment models to simulate the propagation of radar signals through the environment using some combination of ray tracing \cite{coleman1998ray, auer2016raysar, hirsenkorn2017ray, schoffmann2021virtual, schusler2021realistic}, finite element modeling (FEM) \cite{malinen2013elmer, di2004migration}, or finite-difference time domain (FDTD) simulation \cite{furse1990improvements, teixeira1998finite, ernst2007full}. While simulators can replicate complex scene dynamics (e.g. occlusion, path loss, multipath, non-Lambertian effects), they make no attempt to infer the environment structure from sensor data, and their accuracy is limited by the user's ability to create a radar-realistic model of the environment.

\paragraph{Data-Driven Approaches} \textit{Data-driven} methods use real sensor scans to build an environment model. \emph{Sparse} methods use constant false alarm rate detection (CFAR) to detect discrete reflectors in the environment \cite{rohling1983radar, minkler1990cfar, doer2021yaw}. On the other hand, \emph{dense} methods divide the environment into an explicit voxel grid and infer the radar properties of each cell.

Dense methods can be further divided into coherent and incoherent aggregation. If a fixed (e.g. linear or circular) trajectory or sub-wavelength-accurate pose estimates are available, Synthetic Aperture Radar (SAR) can be used \cite{mamandipoor201460,yamada2017high,yanik2019near,prabhakara2020osprey,qian2020,Mostajabi_2020_CVPR_Workshops}; however, this is impractical for a mobile platform over large areas. Instead, sensor readings (with high angular resolution via many antennas or SAR on smaller pieces of a trajectory) can also be aggregated in an \textit{incoherent} manner, which has been referred to as multi-view 3D reconstruction \cite{laviada2017multiview, laviada2017multistatic,laviada2018multiview} and radargrammetry \cite{crosetto2000radargrammetry}.

\subsection{Machine Learning Methods in Radar}

Many classical radar problems such as radar super-resolution \cite{guan2020through, fang2020superrf, geiss2020radar, sun2021deeppoint, cheng2022novel, prabhakara2022high, gao2021mimo, prabhakara2022exploring}, odometry \cite{lu2020milliego, almalioglu2020milli}, mapping \cite{lu2020see}, activity recognition \cite{singh2019radhar, lien2016soli, wang2021m, xue2022m4esh}, and object classification \cite{kosuge2022mmwave, zhao2020point, shuai2021millieye} have been applied to cheaper, lighter, and more compact radar systems using machine learning. We now seek to solve the novel view synthesis problem from compact, low resolution radars while implicitly creating a higher resolution map.

\subsection{Neural Radiance Fields}

Instead of defining an \textit{explicit} inverse imaging algorithm that recovers a representation of the scene from sensor readings, Neural Radiance Fields \cite{mildenhall2021nerf} \textit{implicitly} invert a forward rendering function through stochastic gradient descent. This requires the following components:
\begin{enumerate}[noitemsep,topsep=0pt,leftmargin=*]
    \item \textbf{World model}: NeRF defines the world as an RGB color and transparency for each position and viewing angle; subsequent works have generalized this to handle anti-aliasing \cite{barron2021mip}, different cameras, and lighting \cite{martinbrualla2020nerfw, tancik2022block}.

    \item \textbf{World representation}: Beyond neural networks \cite{mildenhall2021nerf} or voxel grids \cite{liu2020neural}, more recent works have explored spatial hash tables \cite{muller2022instant} as well as function decompositions for view angle dependence \cite{fridovich2022plenoxels, yu2021plenoctrees}.

    \item \textbf{Rendering function and Model Inversion}: NeRFs model each pixel as a ray and ray-trace the radiance field. The invertibility of this rendering function is crucial: by assuming that each pixel is a ray, the NeRF is ``supervised'' by one RGB image pixel per ray, allowing NeRF to ``solve'' for the few opaque points along the ray.
\end{enumerate}
We innovate on these key enablers of NeRFs in order to apply this approach to mmWave radars.  By applying NeRF techniques to radar, we hope to leverage the extensive body of neural radiance field literature, while also unlocking the potential of neural-implicit representations.

\paragraph{Beyond Visual Fields}

The success of NeRFs has inspired numerous other efforts to apply the same general principle to other sensors, including spatial audio \cite{luo2022learning}, imaging sonar \cite{qadri2022neural, reed2023neural}, LIDAR simulations \cite{huang2023neural}, and RSSI (Received Signal Strength Indicator) mapping \cite{zhao2023nerf}. NeRFs have also been applied to radar \cite{snyder2023extending, jamora2023utilizing} for camera-like high-resolution Synthetic Aperture Radar instead of the compact and inexpensive radars we explore in this paper.
\section{\name: Doppler-Aided Radar Tomography}

While our overall approach is inspired by Neural Radiance Fields, the physics of radar presents several new challenges. We make the following key design decisions (Fig.~\ref{fig:dart-overview}):
\begin{enumerate}[noitemsep,topsep=0pt,leftmargin=*]
    \item We first choose a radar measurement representation space --- range-Doppler --- that overcomes the poor spatial resolution of compact radars (Sec.~\ref{sec:range-doppler}, \ref{sec:rd-preproc}).
    \item We then choose a model to account for radar-specific effects of electromagnetic wave interaction which are important for realistic view synthesis such as specularity, ghost reflections and partial occlusions (Sec.~\ref{sec:world_model}).
    \item Finally, to effectively train and learn neural implicit maps for radars, we choose a network architecture for an \textit{adaptive grid} world representation, design a range-Doppler \textit{rendering} method, and propose key rendering optimizations (Sec. \ref{sec:world_model} --- \ref{sec:rendering}).
\end{enumerate}

\subsection{Range-Doppler Representation}
\label{sec:range-doppler}

Unlike cameras, radars are active sensors which illuminate a scene by transmitting a radio frequency waveform. Upon processing reflections received from objects in the scene, radars can perceive the world in 3 dimensions --- range, azimuth, and elevation --- as a heatmap indicating the reflectivity of objects at that 3D coordinate \cite{richards2005fundamentals,richards2010principles}.

However, while bulky mechanical radars or large solid-state radar arrays can provide azimuth and elevation resolution close to typical cameras, modern inexpensive and compact solid-state radars feature small antenna arrays which make them far inferior in the azimuth and elevation axes \cite{iovescu2017fundamentals}. As a result, these compact radars can only generate coarse heatmaps (\textgreater$15^\circ$ resolution) in the azimuth and elevation axes, causing each range-azimuth-elevation bin to point to a coarse region of 3D space which is far less sharp than a ray from a camera pixel \cite{Long_2021_CVPR,Li_2023_CVPR,ti:AWR1843AOP}.

\begin{figure}
\centering

\centering
\includegraphics[width=0.5\columnwidth]{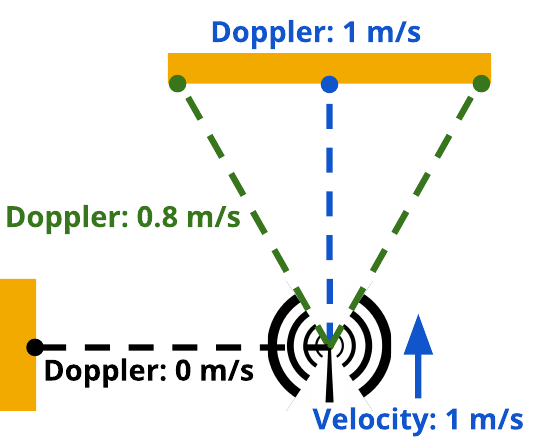}
\hspace{1em}
\includegraphics[width=0.35\columnwidth]{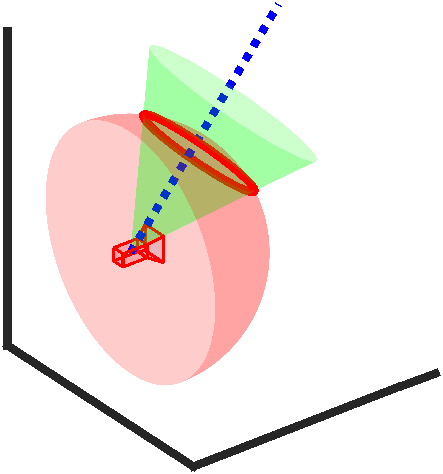}
\vspace{-0.5em}
\caption{Doppler arises due to differences in relative velocities between points with different relative angles to the radar (left). Each range value (red) corresponds to a sphere, while each Doppler value corresponds to a cone (green). The intersection forms the range-Doppler pixel (see Fig. ~\ref{fig:toy_rangedoppler}). }
\vspace{-0.5em}
\label{fig:toy_rowcol}
\end{figure}

To achieve better angular resolution, radars can instead leverage the Doppler effect: objects moving at different relative velocities to the radar have different Doppler velocities, 
which can be measured by examining the residual phase of the range-azimuth-elevation heatmap \cite{8751657}. Crucially, in a static scene, these relative velocities depend on not just the relative speed between the radar and the world, but also the relative azimuth and elevation angle between objects and the radar, with each Doppler corresponding to a cone in space \cite{richards2010principles}. Because of the fine range and Doppler resolutions, Doppler greatly reduces the ambiguity of each bin in 3D space down to a thin ring (Fig.~\ref{fig:toy_rowcol}), which we further reduce by making a \textit{thinness} argument across the range and Doppler axis in order to simplify integration down to a circle for radar rendering (Sec.~\ref{sec:rendering}).


\subsection{Radar Pre-Processing}
\label{sec:rd-preproc}

mmWave radars use a waveform called Frequency Modulated Continuous Wave (FMCW), and measure a continuous time signal; we then convert these signals into range-Doppler-antenna heatmaps. To summarize key points of our radar processing pipeline (Appendix~\ref{appendix:processing}):
\begin{itemize}
    \item \textbf{Undesirable Range-Doppler Side Lobes}: A single reflective object can create sidelobes that bleed into several range-Doppler bins and mask off weaker objects \cite{richards2005fundamentals,s22114208}. Rather than forcing \name\ to model this, we use a Hann weighting window along both range and Doppler axis to mitigate this effect (Appendix ~\ref{appendix:hann}). 
    \item \textbf{Multiple Antennas}: We perform range-Doppler processing on each of the eight transmit-receive (TX/RX) pairs in our radar. During our rendering process (Sec.~\ref{sec:rendering}), we apply the antenna gain and array factor for each TX/RX pair (Fig.~\ref{fig:dart-overview}), emphasizing 8 sections of the field of view. While our sense of high-quality azimuth-elevation information still stems from leveraging Doppler, this provides some coarse directional information.
\end{itemize}

\subsection{\name's World Model} \label{sec:world_model}

If we had an accurate model of the world and the electromagnetic wave interaction for all objects in the world, we could just apply this model to the region defined by each range-Doppler pixel to calculate its value. However, due to the complex nature of real-world scenes and interactions, both tasks are highly difficult and typically impractical. Instead, we model these properties in a data-driven way, representing the reflectance and transmittance using a view-dependent neural network-based approach.

\paragraph{Modeling RF Reflectivity} Modeling mmWave material interactions is one of the most challenging factors of radar view synthesis. From the perspective of radar, points in space have two key properties: reflectance (the proportion of energy that reflects back), and transmittance (the proportion of energy that continues past) \cite{richards2010principles}. However, millimeter waves also interact with objects differently depending on incidence angles \cite{bansal2020pointillism}; for example, metal surfaces can be specular and may be invisible from certain view points. As such, we model each physical point with a reflectance and transmittance value, each of which depend on the incident angle. We formalize this as
\begin{align}
    \sigma: \R^6 \longrightarrow \R, \quad \quad \alpha: \R^6 \longrightarrow [0, 1],
\end{align}
which model the reflectance $\sigma$ and transmittance $\alpha$ as a function of the position ($\R^3$) and incident angle ($\R^3$) of an incoming radar wave, and allows \name\ to model a wide range of radar phenomena such as partial occlusions, specularity, and ghost reflections (Appendix~\ref{appendix:representation}).

\paragraph{World Representation}


While voxel-based approaches are highly effective for learning visual radiance fields \cite{fridovich2022plenoxels, yu2021plenoctrees}, radar images have a much poorer elevation and azimuth resolution compared to cameras even after exploiting the Doppler axis. This magnifies the difference in spatial resolution that $\sigma$ and $\alpha$ can be resolved for between close and far ranges. Moreover, unlike cameras, our angular resolution is variable at all scales --- be it at a trajectory level, frame-to-frame level or even within a frame (Sec.~\ref{sec:range-doppler}). Similar to NeRFs \cite{mildenhall2021nerf}, we turn to neural implicit representations as a means of creating an ``adaptive'' grid, and base our model on the Instant Neural Graphics Primitive\footnote{
    \cite{muller2022instant} implicitly creates an adaptive world grid by using many spatial hash tables with geometrically increasing resolutions, and resolves the output with a small neural network; we use the same general architecture.
} \cite{muller2022instant}. 

Unlike most visual NeRFs, we do not provide the incident angle as an input to the neural network \cite{tancik2023nerfstudio}. Instead, our architecture (visualized in the center block of Fig.~\ref{fig:dart-overview}) outputs a ``base'' reflectance $\bar{\sigma}$ and transmittance $\bar{\alpha}$, as well as shared spherical harmonics coefficients \cite{yu2021plenoctrees} which are applied to the incident angle as an inner product. In addition to computational advantages, this allows us to directly interpret $(\bar{\sigma}, \bar{\alpha})$ as spherical integrals of our learned reflectance and transmittance functions (Appendix~\ref{appendix:sh}).

We also find that the output activation function on $\sigma$ and $\alpha$ is critical for numerical stability and performance. Since $\sigma$ is unbounded\footnote{
    $\sigma$ can be negative; however, since the observed range-Doppler-antenna heatmaps cannot be negative, $\sigma < 0$ will always increase both train and validation loss, so allowing this does not cause overfitting.
}, we apply a linear activation to $\sigma$. Then, to constrain $\alpha \in [0, 1]$, we apply the activation function
\begin{align}
    f(\alpha) = \exp(\max(0, \alpha)),
\end{align}
which we pair with a custom gradient estimator to handle initialization instability (Appendix~\ref{appendix:alpha}).

\subsection{Radar Rendering and Model Training}
\label{sec:rendering}

We train $\sigma$ and $\alpha$ using a differentiable mapping which generates a multi-antenna range-Doppler heatmap from a given $(\sigma, \alpha)$ network; we refer to this as \textit{radar rendering}. Unlike visual NeRFs, \name\ must account for a range of physical effects in addition to occlusion including path attenuation, antenna gain patterns, and the radar-specific Doppler axis.

\paragraph{Ray Tracing} Consider a single ``ray'' emitted from a radar at position $\bm{x}$ and orientation (rotation matrix) $\bm{A}$ at an incidence angle $\bm{w}$. As the ray travels through space up to the maximum range of the processed (range, Doppler, antenna) image, each point $\bm{x} + r_i\bm{w}$ at range $r_i$ receives a signal of amplitude $u_i$, which is attenuated by a factor of $r_i$ due to free space path loss. Each point then reflects a signal of amplitude $u_i\sigma(\bm{t}_i)$ back towards the radar, and propagates an amplitude of $u_i \alpha(\bm{t}_i)$ onwards. As reflected signals return to the radar, the signal loses an additional attenuation factor of $r_i$, while also suffering from occlusion from $\forall j < i: \alpha(\bm{t}_j)$.

Sampling $r_1 \ldots r_{N_r}$ discretely along the range bins of the processed heatmap across antennas, the radar return amplitude $C(i, k, \bm{w})$ for ray $\bm{w}$ at range bin $i$ and antenna $k$ is
\begin{align}
    C(i, k, \bm{w})
    = g_k(\bm{A}^{-1}\bm{w})
    \frac{\sigma(\bm{x} + r_i\bm{w})}{r_i^2}
    \prod_{i'=1}^{i-1} \alpha(\bm{t}_{i'})^2,
    \label{eq:raytracing}
\end{align}
where $g_k(\bm{A}^{-1}\bm{w})$ is the antenna beamforming gain antenna $k$ at angle $\bm{w}$ (specified relative to the radar orientation $\bm{A}$).

\paragraph{Doppler Integration} For a given pose with radar position $\bm{x}$, velocity $\bm{v}$, and orientation $\bm{A}$, we evaluate the return $\bm{Y}(r_i, d_j, k) \in \R$ at each range-Doppler-antenna bin ($r_i, d_j, k$), synthesizing a view-specific, multi-antenna range-Doppler heatmap. Since the doppler velocity is measured as $d_j = \langle \bm{w}, \bm{v} \rangle$, we integrate the return $C$ along the thin ring corresponding to each bin as:
\begin{align}
    \bm{Y}(r_i, d_j, k)
    \propto \frac{r_i}{||\bm{v}||_2} \int \displaylimits_{\langle \bm{w}, \bm{v} \rangle = d_j, \ ||\bm{w}||_2 = 1}
    \mkern-36mu C(i, k, \bm{w}) \ d\bm{w}
    \label{eq:rendering-integral}
\end{align}
Note that we need to correct for the varying width of the discrete bins as a function of range and radar speed by dividing by the speed $||\bm{v}||_2$ and multiply by $r_i$ (Appendix~\ref{appendix:rendering-equation}).

\begin{figure*}[t]
\centering
\includegraphics[width=\textwidth]{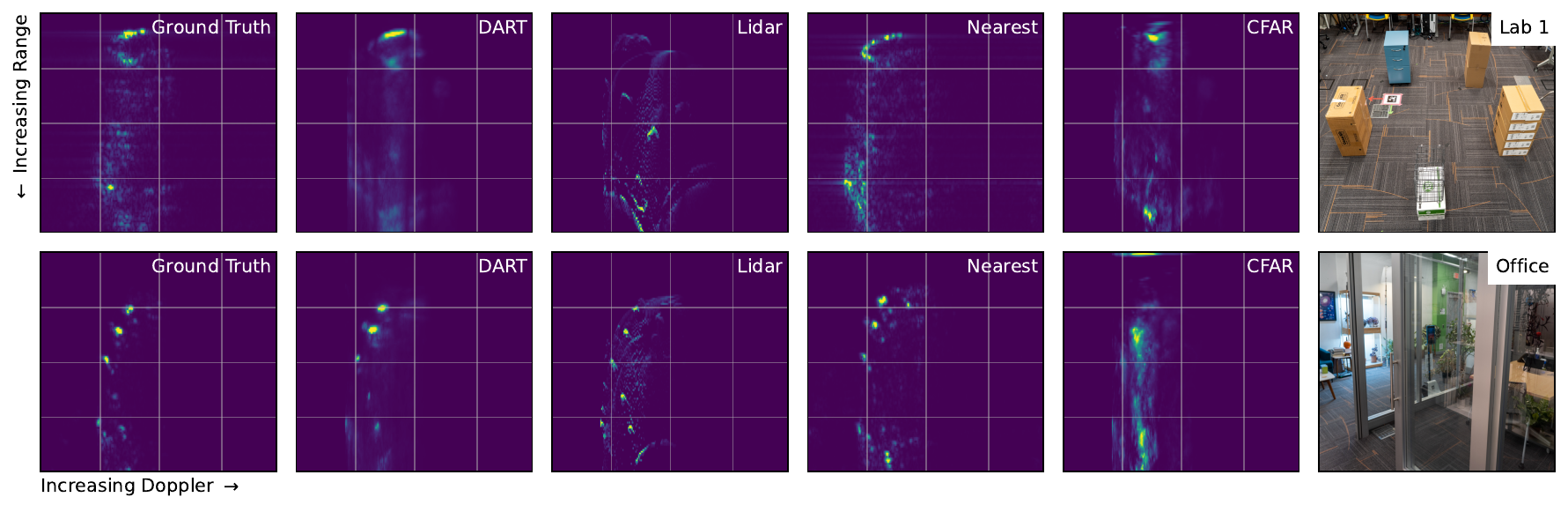}
\vspace*{-2.5em}
\caption{Example (validation) range-Doppler frames and descriptive photos of our method and baselines. \name\ accurately reproduces the overall radar image, though it lacks the resolution to resolve smaller weak reflectors. \textit{Lidar} can model weak reflectors, but cannot accurately scale them due to a lack of radar-specific information, while \textit{Nearest} produces radar-realistic but inaccurate images since exhaustively measuring all possible poses is impractical. Finally, \textit{CFAR} cannot model transmittance or measure the ``volume'' of a point.
}
\vspace{-0.5em}
\label{fig:examples}
\end{figure*}

Approximating this integral as a sum over $M$ random directions $\bm{w}_1 \ldots \bm{w}_M$ such that $\langle \bm{w}, \bm{v} \rangle = d_j$, we multiply by an additional factor of $r_i$ to correct for the circumference of the range-Doppler intersection as $r_i$ increases. This yields
\begin{align}
    \bm{Y}(r_i, d_j, k)
    \propto \frac{r_i^2}{M ||\bm{v}||_2} \sum_{m=1}^M
    C(i, k, \bm{w_m}).
    \label{eq:doppler-integration}
\end{align}

\paragraph{Optimized Rendering}

As the $(\sigma, \alpha)$ field function must be evaluated for every sample, efficient sampling is critical to computational efficiency. Thus, a naive approach of treating each (range, Doppler, antenna) ``pixel'' as an independent sample as is standard practice in NeRFs would be computationally prohibitive, requiring the field to be sampled (range, Doppler, antenna, range integration, Doppler integration) times to render a single image. As such, we aggressively reuse samples of $\sigma$ and $\alpha$ by rendering all bins with the same Doppler simultaneously (Appendix~\ref{appendix:rendering}).

\paragraph{Training}

We train our ($\sigma, \alpha$) field function using stochastic gradient descent with the Adam \cite{kingma2014adam} optimizer and a $l$1 (i.e. mean-absolute-error) loss. For details about our training process and other hyperparameters, see Appendix~\ref{appendix:training}.

\section{Experiments}
\label{sec:experiments}

\begin{figure}
    \centering
    \includegraphics[ width=0.65\columnwidth]{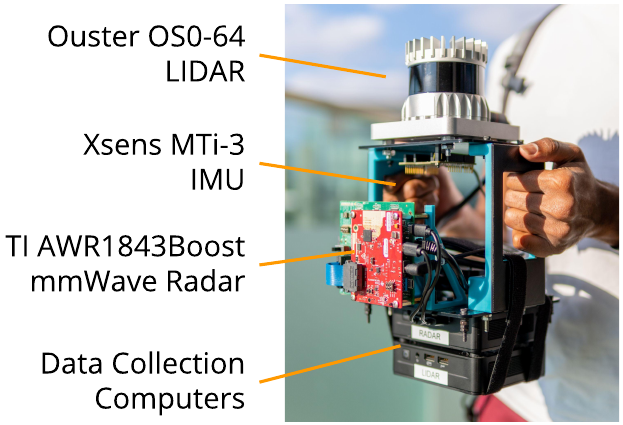}
    \vspace{-0.5em}
    \caption{Handheld data collection system; see Appendix~\ref{appendix:system}.}
    \vspace{-0.5em}
    \label{fig:data-collection-rig}
\end{figure}

We constructed a handheld data collection rig with a mmWave radar and a lidar used for localization\footnote{Note that while we use lidar to obtain pose estimates using Cartographer \cite{hess2016real}, any accurate 3D SLAM system can be used.} (Fig.~\ref{fig:data-collection-rig}; Appendix~\ref{appendix:system}). We used this to collect 12 traces ranging from 5 to 15 minutes long in a diverse set of environments including a lab space, townhouse, high-rise apartment, and an early 20th century house (Appendix~\ref{appendix:datasets}).

\subsection{Baselines}

We implement three baselines for radar novel view synthesis and mapping, a \emph{model-based} approach and two \emph{data-driven} approaches (see Sec.~\ref{sec:radar-simulation}). 
\begin{itemize}
    \item \textbf{Lidar Scan-Based Simulator:} We use lidar scans to create an occupancy grid, which we then use in a raytracing radar simulator (assuming occupied grids have a fixed constant reflectance and no transmittance, similar to \cite{auer2016raysar} without material annotations). This baseline represents the standard practice in radar simulation \cite{auer2016raysar, teixeira1998finite,di2004migration,schoffmann2021virtual}.
    \item \textbf{Nearest Neighbor:} We implement a naive nearest-neighbor baseline which finds the training point with the closest (position, velocity) to the novel viewpoint. While simple, this has the advantage of using radar data to ``simulate'' images compared to our lidar-based simulator \cite{bhatia2010survey}.
    \item \textbf{CFAR Point Cloud Aggregation:} CFAR is a commonly used adaptive algorithm in radar systems to detect target returns against a background of noise, clutter and interference  \cite{minkler1990cfar}. We use the Matlab Phased Array System Toolbox~\cite{cfarmatlab} to detect radar-reflective targets, de-project those targets into 3D points using Bartlett direction-of-arrival estimation \cite{bartlett1950periodogram}, then reproject the points according to the novel pose. This approach is similar to our lidar baseline in that it uses point cloud aggregation, but is better able to capture radar-specific scene properties.
\end{itemize}
For additional details on our baselines, see Appendix~\ref{appendix:baselines}.

\begin{figure*}[t]
\centering
\includegraphics[width=\textwidth]{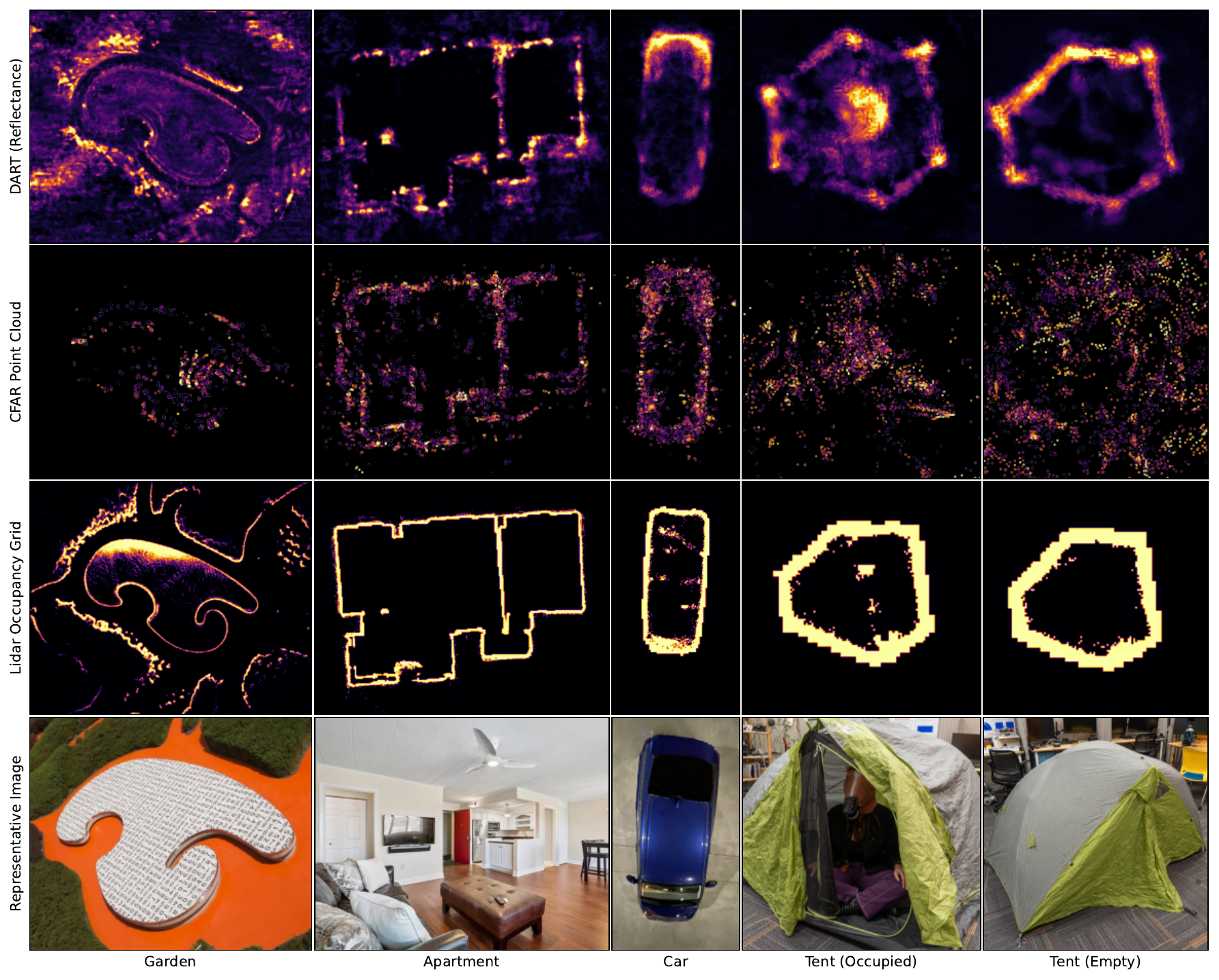}
\vspace{-2em}
\caption{Comparison of \name\ (top) with CFAR (middle) and a Lidar occupancy grid for reference. While CFAR struggles with cluttered scenes and creates point clouds which are both noisy and sparse, \name\ creates relatively clear maps which capture radar-specific properties on both ourdoors (Garden) and indoors (Apartment) environments. \name\ can also image objects with relative clarity (Car), including resolving objects partially occluded by radar-transparent surfaces (Tent --- Occupied/Empty).}
\label{fig:tomography}
\end{figure*}

\subsection{Metrics}

We apply our model to a holdout test set consisting of the last 20\% of each trace. We then compute the SSIM \cite{wang2004image} of the test images and the effective sample size-corrected standard error (SE) for the mean SSIM; see Appendix~\ref{appendix:metrics}. We also compute the SSIM values of 25/30/35dB-equivalent Gaussian noise to help quantify our SSIM values.

\section{Results}
\label{sec:results}

\name\ synthesizes significantly more accurate radar images than each baseline on all traces collected, while using minimal training. We also demonstrate \name's ability to sample tomographic images from its implicit map which are more dense than CFAR point clouds, and more faithfully reproduce radar characteristics than lidar scans.

\paragraph{Training Time}

We train \name\ for 3 epochs on each dataset using a RTX 4090 GPU, taking between 1-2$\times$ the data collection time ($\approx$ 10 minutes) of each dataset\footnote{
    Training time is not directly proportional to the dataset length: since Doppler bins are not observed when the radar speed is less than the Doppler velocity of that bin, we omit these bins, decreasing the training time. See Appendix~\ref{appendix:datasets} for the length and training time of each dataset.
}; this indicates the potential of real-time training with future algorithmic and computing hardware improvements.

\paragraph{Ablations}

Each part of \name's design significantly improves its accuracy, including view dependence using spherical harmonics and our dynamic grid representation (Tab.~\ref{tab:mean_ssim}). For additional ablations, see Appendix~\ref{appendix:ablations}.

\subsection{Comparison with Baselines}

\name\ synthesizes far more accurate radar images than each baseline on all traces in our dataset (Appendix~\ref{appendix:ssim_full}), with the Lidar-based simulator and Nearest Neighbor baselines performing the worst, and CFAR-based simulation in between. \name\ is also significantly better than each baseline when evaluated as a whole (Tab.~\ref{tab:mean_ssim}).

To understand the performance differences between \name\ and each baseline, we selected two example range-Doppler images from our dataset (Fig.~\ref{fig:examples}):
\begin{itemize}
    \item \textit{Lidar-based simulation} (Lidar) can accurately identify reflector positions, but cannot correctly scale their radar return due to the lack of radar-specific material properties.
    \item \textit{Nearest-Neighbor} (Nearest) approaches can, by definition, generate radar-realistic images. However, measuring all possible (position, orientation, velocity) poses is impractical, leading to ``misplaced'' images which do not vary continuously over different poses.
    \item \textit{Constant False Alarm Rate} (CFAR) is commonly used to generate point clouds from radar images. Compared to lidar point clouds, CFAR point clouds are sparse and low-resolution, but capture radar specific properties not measured in lidar. However, CFAR cannot provide any notion of the \textit{size} of each point or its transmittance, which requires the point or grid size to be manually tuned, leading to either excessively sparse or blurry images.
\end{itemize}
\name\ therefore achieves its efficacy by using a domain-appropriate sensor and carefully selecting a representation which allows it to use all available sensor information.

\begin{table}
\small
\centering
\begin{tabular}{l l l}
\toprule
Method & Mean SSIM & SSIM Improvement \\
\toprule
\textbf{\name} & 0.636 $\pm$ 0.012 & --- \\
Lidar & 0.463 $\pm$ 0.005 & 0.174 $\pm$ 0.013 \\
Nearest & 0.468 $\pm$ 0.006 & 0.168 $\pm$ 0.012 \\
CFAR & 0.545 $\pm$ 0.007 & 0.091 $\pm$ 0.006 \\
\hline
No View Dep. & 0.614 $\pm$ 0.015 & 0.022 $\pm$ 0.005 \\
20cm Grid & 0.591 $\pm$ 0.015 & 0.046 $\pm$ 0.004 \\
\toprule
\end{tabular}
\vspace{-0.5em}
\caption{Mean SSIM and SSIM improvement of \name\ over each baseline (and select ablations) across our dataset along with 95\% confidence intervals; see Appendix~\ref{appendix:ssim_full} for a breakdown by dataset.}
\vspace{-0.5em}
\label{tab:mean_ssim}
\end{table}

\subsection{Tomography and Mapping}

\begin{figure}
\centering
\includegraphics[width=\columnwidth]{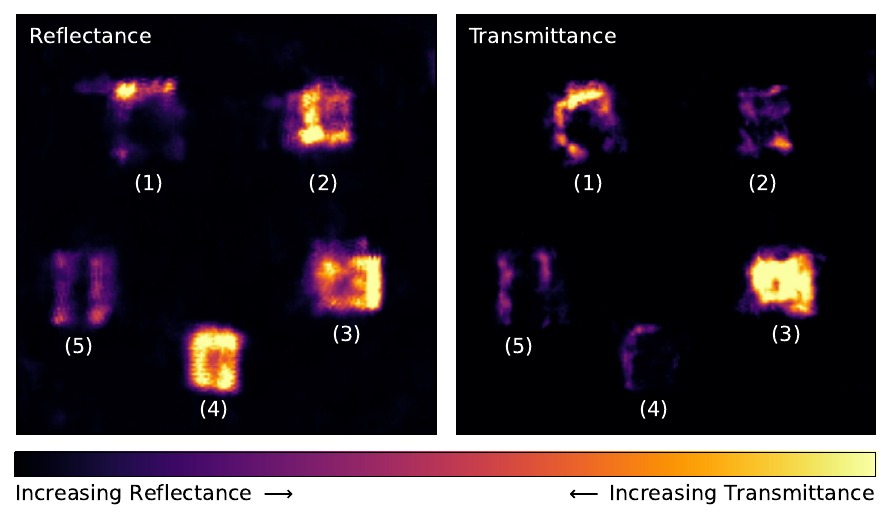}
\includegraphics[width=\columnwidth]{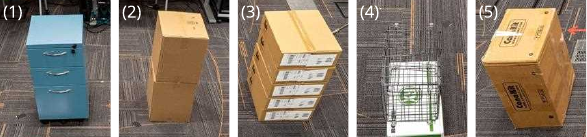}
\vspace{-1.5em}
\caption{Tomographic images of 5 boxes made from different materials: (1) a metal filing cabinet which appears less reflective (due to specularity), but blocks radar waves; (2) an empty box which reflects radar waves but does not block them; (3) a stack of boxes containing electronics equipment which both reflect and block radar waves; (4) a highly reflective metal mesh with large holes that allow radar to penetrate it; and (5) a different empty box which neither reflects nor blocks radar waves.}
\vspace{-0.5em}
\label{fig:boxes}
\end{figure}

While \name\ is not designed primarily as an \textit{explicit} tomography or mapping tool, we can sample the implicit representation\footnote{
    To address view dependence, we analytically take the spherical integral of $\sigma$ and $\alpha$ at each point; see Appendix~\ref{appendix:sh} for details.
} to create a $(\sigma, \alpha)$ reflectance and transmittance grid. This also allows us to verify that \name\ truly learns the mmWave properties of a scene (and does not simply memorize and interpolate the training data).

\paragraph{Material Properties Example} We created an evaluation scene with 5 different boxes. \name\ is able to learn the unique reflectance and transmittance properties of each materials, which we visualize through tomographic reflectance and transmittance maps (Fig.~\ref{fig:boxes}). For additional examples from our datasets, see Appendix~\ref{appendix:tomography}.

\paragraph{Comparison with Baselines} In addition to creating more accurate radar simulations, \name\ can also produce more accurate and dense maps than CFAR. Fig.~\ref{fig:tomography} shows several examples comparing tomographic maps of reflectance learned by \name\ with corresponding slices of the point cloud generated by CFAR. While not as sharp as lidar scans, \name\ produces reasonably clear maps which capture the radar-specific properties of each scene.

\section{Conclusion}

We present DART: Doppler Aided Radar Tomography, a NeRF-inspired radar novel view synthesis algorithm which learns an implicit tomographic map from range-Doppler images, and demonstrate its effectiveness against state-of-the-art baselines. We derive a physics-based rendering model for radar from first principles, and construct an end-to-end system for learning an implicit scene representation and generate realistic novel radar views. While DART provides a strong baseline for future work, many opportunities remain to apply lessons learned from visual NeRFs; given the rapid pace of innovation in NeRF, these opportunities will likely multiply in the coming years. We also currently make a number of assumptions -- such as a static scene and the availability of accurate ground-truth pose -- which could be relaxed as has been done with visual NeRFs, enabling a single-chip radar solution for localization, mapping, and imaging. Finally, as we add mmWave radar to the repertoire of NeRF-enabled sensing technologies, this furthers the potential for multimodal implicit mapping in the future.

{
    \small
    \bibliographystyle{ieeenat_fullname}
    \bibliography{ref}

\begin{thebibliography}{86}
\providecommand{\natexlab}[1]{#1}
\providecommand{\url}[1]{\texttt{#1}}
\expandafter\ifx\csname urlstyle\endcsname\relax
  \providecommand{\doi}[1]{doi: #1}\else
  \providecommand{\doi}{doi: \begingroup \urlstyle{rm}\Url}\fi

\bibitem[cfa()]{cfarmatlab}
Constant false alarm rate (cfar) detection.
\newblock Url: https://www.mathworks.com/help/phased/ug/constant-false-alarm-rate-cfar-detection.html; Accessed: 2023-11-15.

\bibitem[Almalioglu et~al.(2020)Almalioglu, Turan, Lu, Trigoni, and Markham]{almalioglu2020milli}
Yasin Almalioglu, Mehmet Turan, Chris~Xiaoxuan Lu, Niki Trigoni, and Andrew Markham.
\newblock Milli-rio: Ego-motion estimation with low-cost millimetre-wave radar.
\newblock \emph{IEEE Sensors Journal}, 21\penalty0 (3):\penalty0 3314--3323, 2020.

\bibitem[Auer et~al.(2016)Auer, Bamler, and Reinartz]{auer2016raysar}
Stefan Auer, Richard Bamler, and Peter Reinartz.
\newblock Raysar - 3d sar simulator: Now open source.
\newblock 2016.

\bibitem[Bansal et~al.(2020)Bansal, Rungta, Zhu, and Bharadia]{bansal2020pointillism}
Kshitiz Bansal, Keshav Rungta, Siyuan Zhu, and Dinesh Bharadia.
\newblock Pointillism: Accurate 3d bounding box estimation with multi-radars.
\newblock In \emph{Proceedings of the 18th Conference on Embedded Networked Sensor Systems}, pages 340--353, 2020.

\bibitem[Barron et~al.(2021)Barron, Mildenhall, Tancik, Hedman, Martin-Brualla, and Srinivasan]{barron2021mip}
Jonathan~T Barron, Ben Mildenhall, Matthew Tancik, Peter Hedman, Ricardo Martin-Brualla, and Pratul~P Srinivasan.
\newblock Mip-nerf: A multiscale representation for anti-aliasing neural radiance fields.
\newblock In \emph{Proceedings of the IEEE/CVF International Conference on Computer Vision}, pages 5855--5864, 2021.

\bibitem[Bartlett(1950)]{bartlett1950periodogram}
M.~S. Bartlett.
\newblock Periodogram analysis and continuous spectra.
\newblock \emph{Biometrika}, 37\penalty0 (1/2):\penalty0 1--16, 1950.

\bibitem[Bhatia et~al.(2010)]{bhatia2010survey}
Nitin Bhatia et~al.
\newblock Survey of nearest neighbor techniques.
\newblock \emph{arXiv preprint arXiv:1007.0085}, 2010.

\bibitem[Bradbury et~al.(2018)Bradbury, Frostig, Hawkins, Johnson, Leary, Maclaurin, Necula, Paszke, Vander{P}las, Wanderman-{M}ilne, and Zhang]{jax2018github}
James Bradbury, Roy Frostig, Peter Hawkins, Matthew~James Johnson, Chris Leary, Dougal Maclaurin, George Necula, Adam Paszke, Jake Vander{P}las, Skye Wanderman-{M}ilne, and Qiao Zhang.
\newblock {JAX}: composable transformations of {P}ython+{N}um{P}y programs, 2018.

\bibitem[Cen and Newman(2018)]{cen2018precise}
Sarah~H Cen and Paul Newman.
\newblock Precise ego-motion estimation with millimeter-wave radar under diverse and challenging conditions.
\newblock In \emph{2018 IEEE International Conference on Robotics and Automation (ICRA)}, pages 6045--6052. IEEE, 2018.

\bibitem[Cheng et~al.(2022)Cheng, Su, Jiang, and Liu]{cheng2022novel}
Yuwei Cheng, Jingran Su, Mengxin Jiang, and Yimin Liu.
\newblock A novel radar point cloud generation method for robot environment perception.
\newblock \emph{IEEE Transactions on Robotics}, 2022.

\bibitem[Coleman(1998)]{coleman1998ray}
C.~J. Coleman.
\newblock A ray tracing formulation and its application to some problems in over-the-horizon radar.
\newblock \emph{Radio Science}, 33\penalty0 (4):\penalty0 1187--1197, 1998.

\bibitem[Crosetto and P{\'e}rez~Aragues(2000)]{crosetto2000radargrammetry}
Michele Crosetto and F P{\'e}rez~Aragues.
\newblock Radargrammetry and sar interferometry for dem generation: validation and data fusion.
\newblock In \emph{SAR workshop: CEOS committee on earth observation satellites}, page 367, 2000.

\bibitem[di and Wang(2004)]{di2004migration}
Qingyun di and Miaoyue Wang.
\newblock Migration of ground-penetrating radar data method with a finite-element and dispersion.
\newblock \emph{Geophysics}, 69, 2004.

\bibitem[Ding et~al.(2023)Ding, Palffy, Gavrila, and Lu]{Ding_2023_CVPR}
Fangqiang Ding, Andras Palffy, Dariu~M. Gavrila, and Chris~Xiaoxuan Lu.
\newblock Hidden gems: 4d radar scene flow learning using cross-modal supervision.
\newblock In \emph{Proceedings of the IEEE/CVF Conference on Computer Vision and Pattern Recognition (CVPR)}, pages 9340--9349, 2023.

\bibitem[Doer and Trommer(2021)]{doer2021yaw}
Christopher Doer and Gert~F. Trommer.
\newblock Yaw aided radar inertial odometry uisng manhattan world assumptions.
\newblock In \emph{2021 28th Saint Petersburg International Conference on Integrated Navigation Systems (ICINS)}, pages 1--10, 2021.

\bibitem[Ernst et~al.(2007)Ernst, Maurer, Green, and Holliger]{ernst2007full}
Jacques~R. Ernst, Hansruedi Maurer, Alan~G. Green, and Klaus Holliger.
\newblock Full-waveform inversion of crosshole radar data based on 2-d finite-difference time-domain solutions of maxwell's equations.
\newblock \emph{IEEE Transactions on Geoscience and Remote Sensing}, 45\penalty0 (9):\penalty0 2807--2828, 2007.

\bibitem[Fang and Nirjon(2020)]{fang2020superrf}
Shiwei Fang and Shahriar Nirjon.
\newblock Superrf: Enhanced 3d rf representation using stationary low-cost mmwave radar.
\newblock In \emph{International Conference on Embedded Wireless Systems and Networks (EWSN)...}, page 120. NIH Public Access, 2020.

\bibitem[Fridovich-Keil et~al.(2022)Fridovich-Keil, Yu, Tancik, Chen, Recht, and Kanazawa]{fridovich2022plenoxels}
Sara Fridovich-Keil, Alex Yu, Matthew Tancik, Qinhong Chen, Benjamin Recht, and Angjoo Kanazawa.
\newblock Plenoxels: Radiance fields without neural networks.
\newblock In \emph{Proceedings of the IEEE/CVF Conference on Computer Vision and Pattern Recognition}, pages 5501--5510, 2022.

\bibitem[Furse et~al.(1990)Furse, Mathur, and Gandhi]{furse1990improvements}
C.M. Furse, S.P. Mathur, and O.P. Gandhi.
\newblock Improvements to the finite-difference time-domain method for calculating the radar cross section of a perfectly conducting target.
\newblock \emph{IEEE Transactions on Microwave Theory and Techniques}, 38\penalty0 (7):\penalty0 919--927, 1990.

\bibitem[Gao et~al.(2021)Gao, Roy, and Xing]{gao2021mimo}
Xiangyu Gao, Sumit Roy, and Guanbin Xing.
\newblock Mimo-sar: A hierarchical high-resolution imaging algorithm for mmwave fmcw radar in autonomous driving.
\newblock \emph{IEEE Transactions on Vehicular Technology}, 70\penalty0 (8):\penalty0 7322--7334, 2021.

\bibitem[Geiss and Hardin(2020)]{geiss2020radar}
Andrew Geiss and Joseph~C Hardin.
\newblock Radar super resolution using a deep convolutional neural network.
\newblock \emph{Journal of Atmospheric and Oceanic Technology}, 37\penalty0 (12):\penalty0 2197--2207, 2020.

\bibitem[Guan et~al.(2020{\natexlab{a}})Guan, Madani, Jog, Gupta, and Hassanieh]{Guan_2020_CVPR}
Junfeng Guan, Sohrab Madani, Suraj Jog, Saurabh Gupta, and Haitham Hassanieh.
\newblock Through fog high-resolution imaging using millimeter wave radar.
\newblock In \emph{IEEE/CVF Conference on Computer Vision and Pattern Recognition (CVPR)}, 2020{\natexlab{a}}.

\bibitem[Guan et~al.(2020{\natexlab{b}})Guan, Madani, Jog, Gupta, and Hassanieh]{guan2020through}
Junfeng Guan, Sohrab Madani, Suraj Jog, Saurabh Gupta, and Haitham Hassanieh.
\newblock Through fog high-resolution imaging using millimeter wave radar.
\newblock In \emph{Proceedings of the IEEE/CVF Conference on Computer Vision and Pattern Recognition}, pages 11464--11473, 2020{\natexlab{b}}.

\bibitem[Hess et~al.(2016)Hess, Kohler, Rapp, and Andor]{hess2016real}
Wolfgang Hess, Damon Kohler, Holger Rapp, and Daniel Andor.
\newblock Real-time loop closure in 2d lidar slam.
\newblock In \emph{2016 IEEE international conference on robotics and automation (ICRA)}, pages 1271--1278. IEEE, 2016.

\bibitem[Hirsenkorn et~al.(2017)Hirsenkorn, Subkowski, Hanke, Schaermann, Rauch, Rasshofer, and Biebl]{hirsenkorn2017ray}
Nils Hirsenkorn, Paul Subkowski, Timo Hanke, Alexander Schaermann, Andreas Rauch, Ralph Rasshofer, and Erwin Biebl.
\newblock A ray launching approach for modeling an fmcw radar system.
\newblock In \emph{2017 18th International Radar Symposium (IRS)}, pages 1--10, 2017.

\bibitem[Hlawatsch(1998)]{hlawatsch1998time}
Franz Hlawatsch.
\newblock \emph{Time-frequency analysis and synthesis of linear signal spaces: time-frequency filters, signal detection and estimation, and Range-Doppler estimation}.
\newblock Springer Science \& Business Media, 1998.

\bibitem[Huang et~al.(2023)Huang, Gojcic, Wang, Williams, Kasten, Fidler, Schindler, and Litany]{huang2023neural}
Shengyu Huang, Zan Gojcic, Zian Wang, Francis Williams, Yoni Kasten, Sanja Fidler, Konrad Schindler, and Or Litany.
\newblock Neural lidar fields for novel view synthesis.
\newblock \emph{arXiv preprint arXiv:2305.01643}, 2023.

\bibitem[Iovescu and Rao(2017)]{iovescu2017fundamentals}
Cesar Iovescu and Sandeep Rao.
\newblock The fundamentals of millimeter wave sensors.
\newblock \emph{Texas Instruments}, pages 1--8, 2017.

\bibitem[Jamora et~al.(2023)Jamora, Green, Talley, and Curry]{jamora2023utilizing}
JR Jamora, Dylan Green, Ander Talley, and Thomas Curry.
\newblock Utilizing sar imagery in three-dimensional neural radiance fields-based applications.
\newblock In \emph{Algorithms for Synthetic Aperture Radar Imagery XXX}, page 1252002. SPIE, 2023.

\bibitem[Khatun et~al.(2017)Khatun, Mehrpouyan, Matolak, and Guvenc]{8102042}
Mahfuza Khatun, Hani Mehrpouyan, David Matolak, and Ismail Guvenc.
\newblock Millimeter wave systems for airports and short-range aviation communications: A survey of the current channel models at mmwave frequencies.
\newblock In \emph{2017 IEEE/AIAA 36th Digital Avionics Systems Conference (DASC)}, pages 1--8, 2017.

\bibitem[Kingma and Ba(2014)]{kingma2014adam}
Diederik~P Kingma and Jimmy Ba.
\newblock Adam: A method for stochastic optimization.
\newblock \emph{arXiv preprint arXiv:1412.6980}, 2014.

\bibitem[Kosuge et~al.(2022)Kosuge, Suehiro, Hamada, and Kuroda]{kosuge2022mmwave}
Atsutake Kosuge, Satoshi Suehiro, Mototsugu Hamada, and Tadahiro Kuroda.
\newblock mmwave-yolo: A mmwave imaging radar-based real-time multiclass object recognition system for adas applications.
\newblock \emph{IEEE Transactions on Instrumentation and Measurement}, 71:\penalty0 1--10, 2022.

\bibitem[Laviada et~al.(2017{\natexlab{a}})Laviada, Arboleya-Arboleya, {\'A}lvarez, Gonz{\'a}lez-Vald{\'e}s, and Las-Heras]{laviada2017multiview}
Jaime Laviada, Ana Arboleya-Arboleya, Yuri {\'A}lvarez, Borja Gonz{\'a}lez-Vald{\'e}s, and Fernando Las-Heras.
\newblock Multiview three-dimensional reconstruction by millimetre-wave portable camera.
\newblock \emph{Scientific reports}, 7\penalty0 (1):\penalty0 6479, 2017{\natexlab{a}}.

\bibitem[Laviada et~al.(2017{\natexlab{b}})Laviada, Arboleya-Arboleya, and Las-Heras]{laviada2017multistatic}
Jaime Laviada, Ana Arboleya-Arboleya, and Fernando Las-Heras.
\newblock Multistatic millimeter-wave imaging by multiview portable camera.
\newblock \emph{IEEE Access}, 5:\penalty0 19259--19268, 2017{\natexlab{b}}.

\bibitem[Laviada et~al.(2018)Laviada, Lopez-Portugues, Arboleya-Arboleya, and Las-Heras]{laviada2018multiview}
Jaime Laviada, Miguel Lopez-Portugues, Ana Arboleya-Arboleya, and Fernando Las-Heras.
\newblock Multiview mm-wave imaging with augmented depth camera information.
\newblock \emph{IEEE Access}, 6:\penalty0 16869--16877, 2018.

\bibitem[Li et~al.(2022{\natexlab{a}})Li, Wang, Berntorp, and Liu]{Li_2022_CVPR_1}
Peizhao Li, Pu Wang, Karl Berntorp, and Hongfu Liu.
\newblock Exploiting temporal relations on radar perception for autonomous driving.
\newblock In \emph{Proceedings of the IEEE/CVF Conference on Computer Vision and Pattern Recognition (CVPR)}, pages 17071--17080, 2022{\natexlab{a}}.

\bibitem[Li et~al.(2022{\natexlab{b}})Li, Park, O'Toole, and Kitani]{Li_2022_CVPR_2}
Yu-Jhe Li, Jinhyung Park, Matthew O'Toole, and Kris Kitani.
\newblock Modality-agnostic learning for radar-lidar fusion in vehicle detection.
\newblock In \emph{Proceedings of the IEEE/CVF Conference on Computer Vision and Pattern Recognition (CVPR)}, pages 918--927, 2022{\natexlab{b}}.

\bibitem[Li et~al.(2023)Li, Hunt, Park, O{\textquoteright}Toole, and Kitani]{Li_2023_CVPR}
Yu-Jhe Li, Shawn Hunt, Jinhyung Park, Matthew O{\textquoteright}Toole, and Kris Kitani.
\newblock Azimuth super-resolution for fmcw radar in autonomous driving.
\newblock In \emph{Proceedings of the IEEE/CVF Conference on Computer Vision and Pattern Recognition (CVPR)}, pages 17504--17513, 2023.

\bibitem[Lien et~al.(2016)Lien, Gillian, Karagozler, Amihood, Schwesig, Olson, Raja, and Poupyrev]{lien2016soli}
Jaime Lien, Nicholas Gillian, M~Emre Karagozler, Patrick Amihood, Carsten Schwesig, Erik Olson, Hakim Raja, and Ivan Poupyrev.
\newblock Soli: Ubiquitous gesture sensing with millimeter wave radar.
\newblock \emph{ACM Transactions on Graphics (TOG)}, 35\penalty0 (4):\penalty0 1--19, 2016.

\bibitem[Liu et~al.(2020)Liu, Gu, Zaw~Lin, Chua, and Theobalt]{liu2020neural}
Lingjie Liu, Jiatao Gu, Kyaw Zaw~Lin, Tat-Seng Chua, and Christian Theobalt.
\newblock Neural sparse voxel fields.
\newblock \emph{Advances in Neural Information Processing Systems}, 33:\penalty0 15651--15663, 2020.

\bibitem[Long et~al.(2021)Long, Morris, Liu, Castro, Chakravarty, and Narayanan]{Long_2021_CVPR}
Yunfei Long, Daniel Morris, Xiaoming Liu, Marcos Castro, Punarjay Chakravarty, and Praveen Narayanan.
\newblock Radar-camera pixel depth association for depth completion.
\newblock In \emph{Proceedings of the IEEE/CVF Conference on Computer Vision and Pattern Recognition (CVPR)}, pages 12507--12516, 2021.

\bibitem[Lu et~al.(2020{\natexlab{a}})Lu, Rosa, Zhao, Wang, Chen, Stankovic, Trigoni, and Markham]{lu2020see}
Chris~Xiaoxuan Lu, Stefano Rosa, Peijun Zhao, Bing Wang, Changhao Chen, John~A Stankovic, Niki Trigoni, and Andrew Markham.
\newblock See through smoke: robust indoor mapping with low-cost mmwave radar.
\newblock In \emph{Proceedings of the 18th International Conference on Mobile Systems, Applications, and Services}, pages 14--27, 2020{\natexlab{a}}.

\bibitem[Lu et~al.(2020{\natexlab{b}})Lu, Saputra, Zhao, Almalioglu, De~Gusmao, Chen, Sun, Trigoni, and Markham]{lu2020milliego}
Chris~Xiaoxuan Lu, Muhamad Risqi~U Saputra, Peijun Zhao, Yasin Almalioglu, Pedro~PB De~Gusmao, Changhao Chen, Ke Sun, Niki Trigoni, and Andrew Markham.
\newblock milliego: single-chip mmwave radar aided egomotion estimation via deep sensor fusion.
\newblock In \emph{Proceedings of the 18th Conference on Embedded Networked Sensor Systems}, pages 109--122, 2020{\natexlab{b}}.

\bibitem[Luo et~al.(2022)Luo, Du, Tarr, Tenenbaum, Torralba, and Gan]{luo2022learning}
Andrew Luo, Yilun Du, Michael Tarr, Josh Tenenbaum, Antonio Torralba, and Chuang Gan.
\newblock Learning neural acoustic fields.
\newblock In \emph{Advances in Neural Information Processing Systems}, pages 3165--3177. Curran Associates, Inc., 2022.

\bibitem[Malinen and Råback(2013)]{malinen2013elmer}
M. Malinen and P. Råback.
\newblock \emph{Elmer finite element solver for multiphysics and multiscale problems}.
\newblock Forschungszentrum Juelich, 2013.

\bibitem[Mamandipoor et~al.(2014)Mamandipoor, Malysa, Arbabian, Madhow, and Noujeim]{mamandipoor201460}
Babak Mamandipoor, Greg Malysa, Amin Arbabian, Upamanyu Madhow, and Karam Noujeim.
\newblock 60 ghz synthetic aperture radar for short-range imaging: Theory and experiments.
\newblock In \emph{2014 48th Asilomar Conference on Signals, Systems and Computers}, pages 553--558. IEEE, 2014.

\bibitem[Martin-Brualla et~al.(2021)Martin-Brualla, Radwan, Sajjadi, Barron, Dosovitskiy, and Duckworth]{martinbrualla2020nerfw}
Ricardo Martin-Brualla, Noha Radwan, Mehdi S.~M. Sajjadi, Jonathan~T. Barron, Alexey Dosovitskiy, and Daniel Duckworth.
\newblock {NeRF in the Wild: Neural Radiance Fields for Unconstrained Photo Collections}.
\newblock In \emph{CVPR}, 2021.

\bibitem[Mildenhall et~al.(2021)Mildenhall, Srinivasan, Tancik, Barron, Ramamoorthi, and Ng]{mildenhall2021nerf}
Ben Mildenhall, Pratul~P Srinivasan, Matthew Tancik, Jonathan~T Barron, Ravi Ramamoorthi, and Ren Ng.
\newblock Nerf: Representing scenes as neural radiance fields for view synthesis.
\newblock \emph{Communications of the ACM}, 65\penalty0 (1):\penalty0 99--106, 2021.

\bibitem[{Minkler} and {Minkler}(1990)]{minkler1990cfar}
G. {Minkler} and J. {Minkler}.
\newblock {CFAR: The principles of automatic radar detection in clutter}.
\newblock \emph{NASA STI/Recon Technical Report A}, 90:\penalty0 23371, 1990.

\bibitem[Mostajabi et~al.(2020)Mostajabi, Wang, Ranjan, and Hsyu]{Mostajabi_2020_CVPR_Workshops}
Mohammadreza Mostajabi, Ching~Ming Wang, Darsh Ranjan, and Gilbert Hsyu.
\newblock High-resolution radar dataset for semi-supervised learning of dynamic objects.
\newblock In \emph{Proceedings of the IEEE/CVF Conference on Computer Vision and Pattern Recognition (CVPR) Workshops}, 2020.

\bibitem[M{\"u}ller et~al.(2022)M{\"u}ller, Evans, Schied, and Keller]{muller2022instant}
Thomas M{\"u}ller, Alex Evans, Christoph Schied, and Alexander Keller.
\newblock Instant neural graphics primitives with a multiresolution hash encoding.
\newblock \emph{ACM Transactions on Graphics (ToG)}, 41\penalty0 (4):\penalty0 1--15, 2022.

\bibitem[Prabhakara et~al.(2020)Prabhakara, Singh, Kumar, and Rowe]{prabhakara2020osprey}
Akarsh Prabhakara, Vaibhav Singh, Swarun Kumar, and Anthony Rowe.
\newblock Osprey: a mmwave approach to tire wear sensing.
\newblock In \emph{Proceedings of the 18th International Conference on Mobile Systems, Applications, and Services}, pages 28--41, 2020.

\bibitem[Prabhakara et~al.(2022)Prabhakara, Zhang, Li, Munir, Sankaranarayanan, Rowe, and Kumar]{prabhakara2022exploring}
Akarsh Prabhakara, Diana Zhang, Chao Li, Sirajum Munir, Aswin~C Sankaranarayanan, Anthony Rowe, and Swarun Kumar.
\newblock Exploring mmwave radar and camera fusion for high-resolution and long-range depth imaging.
\newblock In \emph{2022 IEEE/RSJ International Conference on Intelligent Robots and Systems (IROS)}, pages 3995--4002. IEEE, 2022.

\bibitem[Prabhakara et~al.(2023)Prabhakara, Jin, Das, Bhatt, Kumari, Soltanaghaei, Bilmes, Kumar, and Rowe]{prabhakara2022high}
Akarsh Prabhakara, Tao Jin, Arnav Das, Gantavya Bhatt, Lilly Kumari, Elahe Soltanaghaei, Jeff Bilmes, Swarun Kumar, and Anthony Rowe.
\newblock High resolution point clouds from mmwave radar.
\newblock In \emph{2023 IEEE International Conference on Robotics and Automation (ICRA)}, 2023.

\bibitem[Qadri et~al.(2022)Qadri, Kaess, and Gkioulekas]{qadri2022neural}
Mohamad Qadri, Michael Kaess, and Ioannis Gkioulekas.
\newblock Neural implicit surface reconstruction using imaging sonar.
\newblock \emph{arXiv preprint arXiv:2209.08221}, 2022.

\bibitem[Qian et~al.(2020)Qian, He, and Zhang]{qian2020}
Kun Qian, Zhaoyuan He, and Xinyu Zhang.
\newblock 3d point cloud generation with millimeter-wave radar.
\newblock \emph{Proc. ACM Interact. Mob. Wearable Ubiquitous Technol.}, 4\penalty0 (4), 2020.

\bibitem[Qian et~al.(2021)Qian, Zhu, Zhang, and Li]{Qian_2021_CVPR}
Kun Qian, Shilin Zhu, Xinyu Zhang, and Li~Erran Li.
\newblock Robust multimodal vehicle detection in foggy weather using complementary lidar and radar signals.
\newblock In \emph{Proceedings of the IEEE/CVF Conference on Computer Vision and Pattern Recognition (CVPR)}, pages 444--453, 2021.

\bibitem[Rebut et~al.(2022)Rebut, Ouaknine, Malik, and P\'erez]{Rebut_2022_CVPR}
Julien Rebut, Arthur Ouaknine, Waqas Malik, and Patrick P\'erez.
\newblock Raw high-definition radar for multi-task learning.
\newblock In \emph{Proceedings of the IEEE/CVF Conference on Computer Vision and Pattern Recognition (CVPR)}, pages 17021--17030, 2022.

\bibitem[Reed et~al.(2023)Reed, Kim, Blanford, Pediredla, Brown, and Jayasuriya]{reed2023neural}
Albert~W Reed, Juhyeon Kim, Thomas Blanford, Adithya Pediredla, Daniel~C Brown, and Suren Jayasuriya.
\newblock Neural volumetric reconstruction for coherent synthetic aperture sonar.
\newblock \emph{arXiv preprint arXiv:2306.09909}, 2023.

\bibitem[Richards et~al.(2010)Richards, Scheer, Holm, and Melvin]{richards2010principles}
Mark~A Richards, Jim Scheer, William~A Holm, and William~L Melvin.
\newblock \emph{Principles of Modern Radar}.
\newblock Citeseer, 2010.

\bibitem[Richards et~al.(2005)]{richards2005fundamentals}
Mark~A Richards et~al.
\newblock \emph{Fundamentals of radar signal processing}.
\newblock Mcgraw-hill New York, 2005.

\bibitem[Robert et~al.(1999)Robert, Casella, and Casella]{robert1999monte}
Christian~P Robert, George Casella, and George Casella.
\newblock \emph{Monte Carlo statistical methods}.
\newblock Springer, 1999.

\bibitem[Rohling(1983)]{rohling1983radar}
Hermann Rohling.
\newblock Radar cfar thresholding in clutter and multiple target situations.
\newblock \emph{IEEE Transactions on Aerospace and Electronic Systems}, AES-19\penalty0 (4):\penalty0 608--621, 1983.

\bibitem[Russell et~al.(1997)Russell, Crain, Curran, Campbell, Drubin, and Miccioli]{russell1997millimeter}
Mark~E Russell, Arthur Crain, Anthony Curran, Richard~A Campbell, Clifford~A Drubin, and Willian~F Miccioli.
\newblock Millimeter-wave radar sensor for automotive intelligent cruise control (icc).
\newblock \emph{IEEE Transactions on microwave theory and techniques}, 45\penalty0 (12):\penalty0 2444--2453, 1997.

\bibitem[Scheiner et~al.(2020)Scheiner, Kraus, Wei, Phan, Mannan, Appenrodt, Ritter, Dickmann, Dietmayer, Sick, and Heide]{Scheiner_2020_CVPR}
Nicolas Scheiner, Florian Kraus, Fangyin Wei, Buu Phan, Fahim Mannan, Nils Appenrodt, Werner Ritter, Jurgen Dickmann, Klaus Dietmayer, Bernhard Sick, and Felix Heide.
\newblock Seeing around street corners: Non-line-of-sight detection and tracking in-the-wild using doppler radar.
\newblock In \emph{IEEE/CVF Conference on Computer Vision and Pattern Recognition (CVPR)}, 2020.

\bibitem[Schöffmann et~al.(2021)Schöffmann, Ubezio, Böhm, Mühlbacher-Karrer, and Zangl]{schoffmann2021virtual}
Christian Schöffmann, Barnaba Ubezio, Christoph Böhm, Stephan Mühlbacher-Karrer, and Hubert Zangl.
\newblock Virtual radar: Real-time millimeter-wave radar sensor simulation for perception-driven robotics.
\newblock \emph{IEEE Robotics and Automation Letters}, 6\penalty0 (3):\penalty0 4704--4711, 2021.

\bibitem[Schüßler et~al.(2021)Schüßler, Hoffmann, Bräunig, Ullmann, Ebelt, and Vossiek]{schusler2021realistic}
Christian Schüßler, Marcel Hoffmann, Johanna Bräunig, Ingrid Ullmann, Randolf Ebelt, and Martin Vossiek.
\newblock A realistic radar ray tracing simulator for large mimo-arrays in automotive environments.
\newblock \emph{IEEE Journal of Microwaves}, 1\penalty0 (4):\penalty0 962--974, 2021.

\bibitem[Sheen et~al.(1998)Sheen, McMakin, and Hall]{sheen1998cylindrical}
David~M Sheen, Douglas~L McMakin, and Thomas~E Hall.
\newblock Cylindrical millimeter-wave imaging technique for concealed weapon detection.
\newblock In \emph{26th AIPR Workshop: Exploiting New Image Sources and Sensors}, pages 242--250. SPIE, 1998.

\bibitem[Shuai et~al.(2021)Shuai, Shen, Tang, Shi, Ji, and Xing]{shuai2021millieye}
Xian Shuai, Yulin Shen, Yi Tang, Shuyao Shi, Luping Ji, and Guoliang Xing.
\newblock millieye: A lightweight mmwave radar and camera fusion system for robust object detection.
\newblock In \emph{Proceedings of the International Conference on Internet-of-Things Design and Implementation}, pages 145--157, 2021.

\bibitem[Singh et~al.(2019)Singh, Sandha, Garcia, and Srivastava]{singh2019radhar}
Akash~Deep Singh, Sandeep~Singh Sandha, Luis Garcia, and Mani Srivastava.
\newblock Radhar: Human activity recognition from point clouds generated through a millimeter-wave radar.
\newblock In \emph{Proceedings of the 3rd ACM Workshop on Millimeter-wave Networks and Sensing Systems}, pages 51--56, 2019.

\bibitem[Snyder et~al.(2023)Snyder, DelMarco, Snover, Bhatia, and Kuzdeba]{snyder2023extending}
William Snyder, Stephen DelMarco, Dylan Snover, Amit Bhatia, and Scott Kuzdeba.
\newblock Extending neural radiance fields (nerf) for synthetic aperture radar (sar) novel image generation.
\newblock In \emph{Synthetic Data for Artificial Intelligence and Machine Learning: Tools, Techniques, and Applications}, pages 268--276. SPIE, 2023.

\bibitem[Sun et~al.(2021)Sun, Zhang, Huang, and Liu]{sun2021deeppoint}
Yue Sun, Honggang Zhang, Zhuoming Huang, and Benyuan Liu.
\newblock Deeppoint: A deep learning model for 3d reconstruction in point clouds via mmwave radar.
\newblock \emph{arXiv preprint arXiv:2109.09188}, 2021.

\bibitem[Tancik et~al.(2022)Tancik, Casser, Yan, Pradhan, Mildenhall, Srinivasan, Barron, and Kretzschmar]{tancik2022block}
Matthew Tancik, Vincent Casser, Xinchen Yan, Sabeek Pradhan, Ben Mildenhall, Pratul~P Srinivasan, Jonathan~T Barron, and Henrik Kretzschmar.
\newblock Block-nerf: Scalable large scene neural view synthesis.
\newblock In \emph{Proceedings of the IEEE/CVF Conference on Computer Vision and Pattern Recognition}, pages 8248--8258, 2022.

\bibitem[Tancik et~al.(2023)Tancik, Weber, Ng, Li, Yi, Kerr, Wang, Kristoffersen, Austin, Salahi, et~al.]{tancik2023nerfstudio}
Matthew Tancik, Ethan Weber, Evonne Ng, Ruilong Li, Brent Yi, Justin Kerr, Terrance Wang, Alexander Kristoffersen, Jake Austin, Kamyar Salahi, et~al.
\newblock Nerfstudio: A modular framework for neural radiance field development.
\newblock \emph{arXiv preprint arXiv:2302.04264}, 2023.

\bibitem[Teixeira et~al.(1998)Teixeira, Chew, Straka, Oristaglio, and Wang]{teixeira1998finite}
F.L. Teixeira, Weng~Cho Chew, M. Straka, M.L. Oristaglio, and T. Wang.
\newblock Finite-difference time-domain simulation of ground penetrating radar on dispersive, inhomogeneous, and conductive soils.
\newblock \emph{IEEE Transactions on Geoscience and Remote Sensing}, 36\penalty0 (6):\penalty0 1928--1937, 1998.

\bibitem[Tex(2021)]{ti:AWR1843AOP}
\emph{AWR1843AOP Single-chip 77- and 79-GHz FMCW mmWave Sensor Antennas-OnPackage (AOP)}.
\newblock Texas Instruments, 2021.

\bibitem[Wang et~al.(2021)Wang, Liu, Cui, Zhou, Li, and Ma]{wang2021m}
Yuheng Wang, Haipeng Liu, Kening Cui, Anfu Zhou, Wensheng Li, and Huadong Ma.
\newblock m-activity: Accurate and real-time human activity recognition via millimeter wave radar.
\newblock In \emph{ICASSP 2021-2021 IEEE International Conference on Acoustics, Speech and Signal Processing (ICASSP)}, pages 8298--8302. IEEE, 2021.

\bibitem[Wang et~al.(2004)Wang, Bovik, Sheikh, and Simoncelli]{wang2004image}
Zhou Wang, Alan~C Bovik, Hamid~R Sheikh, and Eero~P Simoncelli.
\newblock Image quality assessment: from error visibility to structural similarity.
\newblock \emph{IEEE transactions on image processing}, 13\penalty0 (4):\penalty0 600--612, 2004.

\bibitem[Wasserzier et~al.(2019)Wasserzier, Worms, and O'Hagan]{8751657}
Christoph Wasserzier, Josef~G. Worms, and Daniel~W. O'Hagan.
\newblock How noise radar technology brings together active sensing and modern electronic warfare techniques in a combined sensor concept.
\newblock In \emph{2019 Sensor Signal Processing for Defence Conference (SSPD)}, pages 1--5, 2019.

\bibitem[Xue et~al.(2022)Xue, Cao, Ju, Hu, Wang, Zhang, and Su]{xue2022m4esh}
Hongfei Xue, Qiming Cao, Yan Ju, Haochen Hu, Haoyu Wang, Aidong Zhang, and Lu Su.
\newblock M4esh: mmwave-based 3d human mesh construction for multiple subjects.
\newblock In \emph{Proceedings of the 20th ACM Conference on Embedded Networked Sensor Systems}, pages 391--406, 2022.

\bibitem[Yamada et~al.(2017)Yamada, Kobayashi, Yamaguchi, and Sugiyama]{yamada2017high}
Hiroyoshi Yamada, Takumi Kobayashi, Yoshio Yamaguchi, and Yuuichi Sugiyama.
\newblock High-resolution 2d sar imaging by the millimeter-wave automobile radar.
\newblock In \emph{2017 IEEE Conference on Antenna Measurements \& Applications (CAMA)}, pages 149--150. IEEE, 2017.

\bibitem[Yanik and Torlak(2019)]{yanik2019near}
Muhammet~Emin Yanik and Murat Torlak.
\newblock Near-field mimo-sar millimeter-wave imaging with sparsely sampled aperture data.
\newblock \emph{Ieee Access}, 7:\penalty0 31801--31819, 2019.

\bibitem[Yu et~al.(2021)Yu, Li, Tancik, Li, Ng, and Kanazawa]{yu2021plenoctrees}
Alex Yu, Ruilong Li, Matthew Tancik, Hao Li, Ren Ng, and Angjoo Kanazawa.
\newblock Plenoctrees for real-time rendering of neural radiance fields.
\newblock In \emph{Proceedings of the IEEE/CVF International Conference on Computer Vision}, pages 5752--5761, 2021.

\bibitem[Zhao et~al.(2023)Zhao, An, Pan, and Yang]{zhao2023nerf}
Xiaopeng Zhao, Zhenlin An, Qingrui Pan, and Lei Yang.
\newblock Nerf2: Neural radio-frequency radiance fields.
\newblock \emph{arXiv preprint arXiv:2305.06118}, 2023.

\bibitem[Zhao et~al.(2020)Zhao, Song, Cui, Zhu, Song, Xu, and Ding]{zhao2020point}
Zihao Zhao, Yuying Song, Fucheng Cui, Jiang Zhu, Chunyi Song, Zhiwei Xu, and Kai Ding.
\newblock Point cloud features-based kernel svm for human-vehicle classification in millimeter wave radar.
\newblock \emph{IEEE Access}, 8:\penalty0 26012--26021, 2020.

\bibitem[Zhou et~al.(2022)Zhou, Liu, Zhao, López-Benítez, Yu, and Yue]{s22114208}
Yi Zhou, Lulu Liu, Haocheng Zhao, Miguel López-Benítez, Limin Yu, and Yutao Yue.
\newblock Towards deep radar perception for autonomous driving: Datasets, methods, and challenges.
\newblock \emph{Sensors}, 22\penalty0 (11), 2022.

\end{thebibliography}
}

\clearpage
\setcounter{page}{1}
\maketitlesupplementary

\appendix
\section{Method Details}

\name\ represents the sum of careful consideration of each step in the novel view synthesis pipeline from radar pre-processing to rendering. We provide additional details not included in our main paper, including our radar processing pipeline (Sec.~\ref{appendix:processing}), spherical harmonics representation (Sec.~\ref{appendix:sh}), transmittance representation and custom gradient estimator (Sec.~\ref{appendix:alpha}), optimized radar rendering pipeline (Sec.~\ref{appendix:rendering}), and hyperparameters (Sec.~\ref{appendix:training}).

\subsection{Radar Pre-Processing}
\label{appendix:processing}

\begin{figure}
    \centering
    \includegraphics[width=\columnwidth]{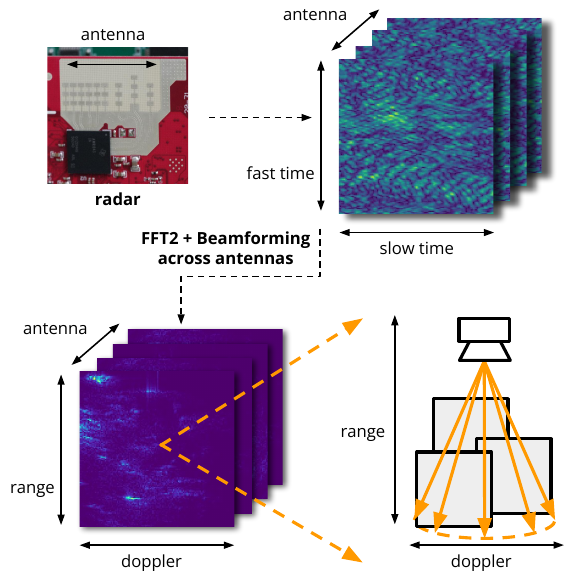}
    \vspace*{-1.5em}
    \caption{\name\ performs range-Doppler pre-processing on each of the antennas in the mmWave radar using 2D FFTs and antenna beamforming gains. A key point to note is the reduction in range-Doppler sidelobes using signal processing windows prior to FFTs for effective \name\ training.}
    \label{fig:radar-processing}
    \vspace{-0.5em}
\end{figure}

Our radar processing pipeline (Fig.~\ref{fig:radar-processing}) consists of a range, Doppler, and azimuth fast Fourier transform (FFT). The range and Doppler FFT use a Hann window in order to suppress undesirable side lobes. Finally, we perform range decimation to take the first 128 valid range bins of each processed range-Doppler-azimuth heatmap, and apply calibration to account for antenna delay.

Note that since we do not have sub-mm pose accuracy or restrict data collection to fixed trajectories, \name\ does not use coherent radar images. As such, while each FFT produces complex values, we only use the magnitude of the final output.

\paragraph{Range FFT}

State-of-the-art solid-state compact radars operate in the radio frequency regime of millimeter wave (mmWave) frequencies (60 GHz / 77 GHz). The wide bandwidths (about 4 GHz) available at these frequencies provide the impressive range resolution of mmWave radars of around 4cm ($\frac{c}{2B}$, where $c$ is speed of light and $B$ is the bandwidth).

This means that the radar can distinguish objects in the world separated by 4 cm apart purely based on range processing. Since the range of an object manifests in the time delay of its reflection, mmWave radars send ``chirps'' in a waveform called Frequency Modulated Continuous Wave (FMCW) which converts time delays to frequency shifts \cite{iovescu2017fundamentals}. We perform range processing by simply computing a 1D Fourier transform of the received reflections which converts the frequency shifts back to time delays that is proportional to range (Range $ R = \frac{c \Delta T}{2}$, where $c$ is speed of light and $\Delta T$ is the two-way time delay).

The time window that this Fourier transformed is performed over is referred to as \textit{fast time}, and is equal to the maximum theoretical number of range bins.

\paragraph{Doppler FFT}

We next compute the Doppler velocities by processing a time series of FMCW waveforms. Unlike range information which is observed as time delay of reflections, Doppler velocities manifest as phase change of the reflected signal when there is relative motion between radar and the world. As the radar moves, an object changes phase as the range $R$ changes slightly over a time window
\begin{align}
\exp \left( \frac{j2\pi2R(t)}{\lambda} \right),
\end{align}
where $\lambda$ is the wavelength corresponding to the FMCW waveform's center frequency. Because of the tiny wavelengths (4 mm) of mmWave signals, even tiny shifts in range show up as significant phase changes. The rate of change of these phase changes over the time series of FMCW waveforms provides the Doppler velocity.

Similar to range processing, we perform another Fourier Transform, but across a time sequence of range processed FMCW reflections. The time window (across different chirps) which this Fourier transform is performed over is referred to as \textit{slow time}.

\paragraph{Impact of Doppler Resolution}

Unlike azimuth or elevation resolution which is governed by the number of antennas in a radar system, the Doppler resolution depends only on the radar frame integration time. In theory, we can have arbitrarily fine Doppler resolution with long frame times; however, this assumes constant radar velocity over the entire integration time window, which is impractical.

Moreover, Doppler resolution can be simplified and mapped to azimuth resolution (in 2D) as 
\begin{align}
    \Delta \theta = \frac{\Delta D \lambda}{2v sin\theta}
\end{align}
Therefore, as $\Delta D$ becomes lower with longer frame integration time, the azimuth resolution becomes better. In our case, the (theoretical) azimuth resolution derived from Doppler is 0.85\textdegree{} for a head-on point at $\theta$=90\textdegree{}, a radar speed of 0.5 m/s and a radar frame integration time of 256 ms --- much better than the best case of 15\textdegree{} that compact radars today are capable of \cite{Rebut_2022_CVPR,ti:AWR1843AOP}.

\paragraph{Hann Filtering}
\label{appendix:hann}

\begin{figure}
\begin{subfigure}[t]{0.48\columnwidth}
    \centering
    \includegraphics[width=\textwidth]{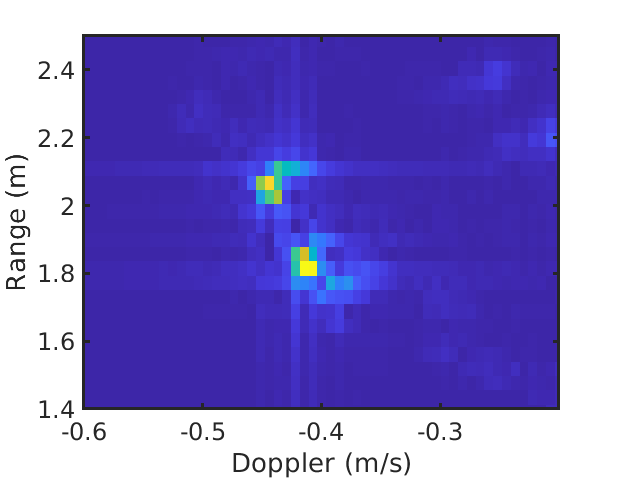}
    \caption{Without Hann Filtering}
\end{subfigure}
\begin{subfigure}[t]{0.48\columnwidth}
    \centering
    \includegraphics[width=\textwidth]{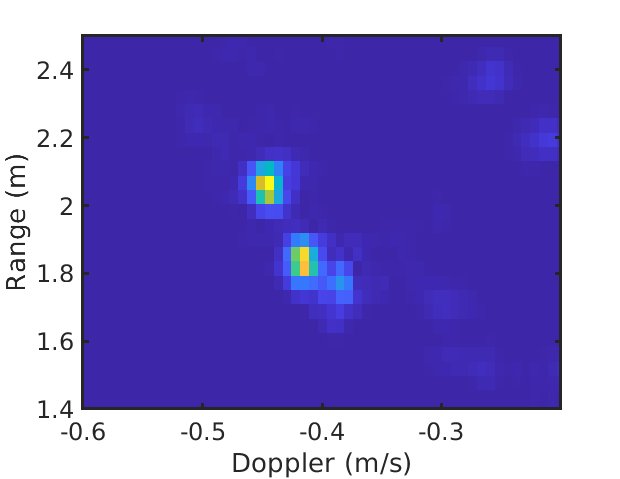}
    \caption{With Hann Filtering}
\end{subfigure}
\vspace{-0.5em}
\caption{Spectral leakage causes inter-bin interference, visible as streaking in the horizontal and vertical directions (range-Doppler images are cropped to a small window). Hann low-pass filtering softens window edges and reduces spectral leakage significantly at the cost of reduced sharpness.}
\label{fig:hann-filtering}
\end{figure}

Since both the range FFT and Doppler FFT are taken over a finite-time window, the hard edges of the time window induce high-frequency effects that result in a phenomenon known as \emph{spectral leakage}. This leakage is a mathematical artifact that can be modeled as a $sinc()$ function and is clearly visible in radar scans \cite{richards2005fundamentals,richards2010principles}. While it would be possible to accurately reproduce such artifacts in DART's forward rendering model, this would require modeling inter-column effects (breaking \name's per-column rendering model) and greatly increase the computational cost of our algorithm. Instead, we apply a Hann window weighting to the raw range-Doppler frames before computing the FFT, which slightly blurs the resulting images but greatly reduces the inter-bin artifacts (Fig. \ref{fig:hann-filtering}).

\paragraph{Azimuth FFT}

After applying the range and Doppler FFT, we also apply an Azimuth FFT across 8 azimuth bins (2 TX $\times$ 4 RX antennas). While this also results in a \textit{sinc} pattern, we model it explicitly in the form of an array factor (Sec.~\ref{appendix:rendering-equation}) applied in addition to the antenna gain on each of the resulting azimuth bins.

\subsection{Reflectance and Transmittance Modeling}
\label{appendix:representation}

\begin{figure}
\centering
\begin{subfigure}[b]{0.156\textwidth}
    \centering
    \includegraphics[width=\textwidth]{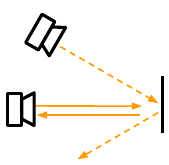}
    \vspace{-1em}\caption{Specularity}
    \label{fig:specularity}
\end{subfigure}
\hspace{0.05\textwidth}
\begin{subfigure}[b]{0.24\textwidth}
    \centering
    \includegraphics[width=\textwidth]{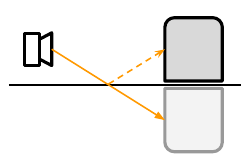}
    \vspace{-1em}\caption{Ghost Reflections}
    \label{fig:reflections}
\end{subfigure}
\vspace{-0.5em}
\caption{View-dependent reflectance and transmittance captures both specularity and ghost reflections.}
\vspace{-0.5em}
\end{figure}

\name's view-dependent transmittance-reflectance model is analagous to the density-color representation of NeRFs, and enables modeling a range of physical phenomena, such as:
\begin{enumerate}[noitemsep,topsep=0pt,leftmargin=*]
    \item Diffuse reflections: This makes up for the majority of first order reflections from non-metallic surfaces.
    \item Partial occlusions: Modeling transmittance allows objects behind surfaces such as drywall to be visible while occluding objects behind opaque surfaces, e.g. concrete.
    \item Specularity (Fig.~\ref{fig:specularity}): This manifests as high 
    reflectance in specific directions, e.g. metallic surfaces.
    \item Ghost reflections (Fig.~\ref{fig:reflections}): This occurs as a result of seeing an object after multiple orders of reflections from other objects, e.g. seeing the ghost reflection of a chair off of the ground when radar is looking at the ground.
\end{enumerate}

\subsection{Spherical Harmonics}
\label{appendix:sh}

Our usage of spherical harmonics provides view-dependence, while also making interpretation of the resulting implicit map easier. In this section, we describe how our usage of spherical harmonics contrasts with common practice in visual Neural Radiance Fields and provides key advantages for our use case.

\paragraph{Spherical Harmonics Representation} Spherical harmonics are a spherical (orthonormal) basis function which is commonly used to decompose functions defined on the unit sphere. Neural Radiance Field models typically handle view dependence by projecting the incidence angle of the incoming ray to between 9 and 25 (real) spherical harmonic coefficients, which is used as an input to the neural network \cite{tancik2023nerfstudio, muller2022instant}, acting as essentially a positional embedding on the viewing angle. We instead use spherical harmonic coefficients as an \textit{output} similar to non-neural-network based radiance field methods \cite{yu2021plenoctrees, fridovich2022plenoxels}.

We use spherical harmonic coefficients of degree 25. Let $\bm{Y}: \R^3 \rightarrow \R^{25}$ be the function that maps incoming incident angles to the degree-25 spherical harmonics. Then, our field function output is given by
\begin{align}
\begin{split}
    \sigma(\bm{w}) &= \bar{\sigma} \langle \bm{Y}(\bm{w}), \bm{c} / ||\bm{c}||_2 \rangle, \\
    \alpha(\bm{w}) &= f\left(\bar{\alpha} \langle \bm{Y}(\bm{w}), \bm{c} / ||\bm{c}||_2 \rangle\right), \\
\end{split}
\end{align}
where $\bar{\sigma} \in \R, \bar{\alpha} \in \R, \bm{c} \in \R^{25}$ are outputs of the neural network, $f$ is the activation function on $\alpha$ (Sec.~\ref{appendix:alpha}), and $\bm{w}$ is the incident angle.

Crucially, unlike visual NeRFs, we also must model transmittance $\alpha$ (which is analogous to the density in NeRF) as having view dependence. While reflective surfaces in visual NeRFs are modeled as ``screens'' whose content changes depending on the viewing angle, radars measure distance, which requires reflective surfaces to instead be measured as ``holes'' which become transparent depending on the viewing angle.

\paragraph{Interpretability} Since spherical harmonics are an orthonormal basis, we can analytically evaluate the spherical $\mathcal{L}_2$ norm of our DART field at each point without requiring numerical integration in order to sample a ``mean'' reflectance and transmittance. For example:
\begin{align}
\begin{split}
    ||\sigma||_2^2 &= \int_w \left(\bar{\sigma} \langle \bm{Y}(\bm{w}), \bm{c} / (||\bm{c}||_2) \rangle \ d\bm{w} \right)^{2} \\
    &= \frac{\bar{\sigma}^2}{||\bm{c}||_2^2} \int_w \bm{c}^T\bm{Y}(\bm{w})\bm{Y}(\bm{w})^T\bm{c} \ d\bm{w} \\
    &= \bar{\sigma}^2
\end{split}
\end{align}
since, by the orthonormality of spherical harmonics $\bm{Y}$,
\begin{align}
    \int_{\bm{w}} \bm{Y}(\bm{w})\bm{Y}(\bm{w})^T \ d\bm{w} = \bm{I},
\end{align}
and neural network output coefficients $\bm{c}$ are unrelated to $\bm{w}$ given a fixed point in space. If we ignore the activation function $f$, the same applies to $\alpha$.

\subsection{Transmittance Representation}
\label{appendix:alpha}

Each ray in our rendering pipeline corresponds to one sampled reflectance $\sigma$ from DART. As such, negative $\sigma$ are immediately penalized by gradient descent, so an activation function on $\sigma$ is not necessary, and can harm the numerical stability of DART (Sec.~\ref{appendix:ablations}).

On the other hand, since each transmittance value $\sigma$ affects all bins behind it, allowing $\alpha > 1$ (i.e. for points in space to transmit \textit{more} energy than they receive) can allow for a significant degree of overfitting. At the same time, we must account for the fact that most points in space are empty (so it should be ``easy'' for the model to represent $\alpha = 1$), while handling initialization instability and the possibility of ``dead'' gradients.

Finally, to further reduce the potential for overfitting, we only allow $\alpha < 1$ when $\sigma$ exceeds a set threshold (0.05 in our experiments). To reduce the ``swiss cheese'' effect caused when this threshold is initialized to a (relatively) large value, we also anneal this threshold in over 600 steps (starting from -1, which in practice is always less than the initialized $\sigma$, to 0.0 linearly in 100 steps, and from 0.0 to 0.05 linearly in 500 steps).

\paragraph{Existing Activation Functions}

No existing activation functions satisfy our desired properties:
\begin{itemize}
    \item Sigmoid: this requires a $-\infty$ input to represent 0, and suffers from vanishing gradients at either end.
    \item Value clipping $\alpha = \text{clip}(\alpha', 0, 1)$: clipping output values to $[0, 1]$ suffers greatly from ``dead'' regions in space, especially where $\alpha$ is initialized at $\alpha' < 0$, and cannot recover.
    \item Partial clipping $\alpha = \exp(\min(0, \alpha'))$: clipping at only one end suffers from the same ``dead'' region problem, though only at one side.
\end{itemize}
Indeed, any activation function on [0, 1] must have vanishing gradients of some form. As such, we instead turn to a custom gradient estimator to create an activation function which satisfies our desired properties.

\paragraph{Gradient Estimator}

Using partial clipping $\alpha = \exp(\min(0, \alpha'))$ as a starting point, we specify our gradient estimator for $\min(0, \alpha')$ as follows:
\begin{itemize}
    \item $\alpha' \leq 0$: gradients pass through, i.e. $d/d\alpha' = d/d\alpha$.
    \item $\alpha' > 0, d/d\alpha' \geq 0$: In this case, $d/d\alpha' > 0$, indicating that the gradients are trying to push $\alpha'$ negative, even though $\min(0, \alpha)$ is ``saturated''. As such, we pass $d/d\alpha' = d/d\alpha$ as well.
    \item $\alpha' > 0, d/d\alpha' < 0$: In this case, the gradients are trying to push $\alpha'$ positive, even though it is already positive. To prevent our parameter values from exploding, we set $d/d\alpha' = 0$.
\end{itemize}

\subsection{Rendering Equation Derivation Details}
\label{appendix:rendering-equation}

\name's rendering process is summarized by Eq.~\ref{eq:doppler-integration} at a high level; in this section, we expand on the derivation of this equation, as well as specific elements of our implementation of the rendering equation.

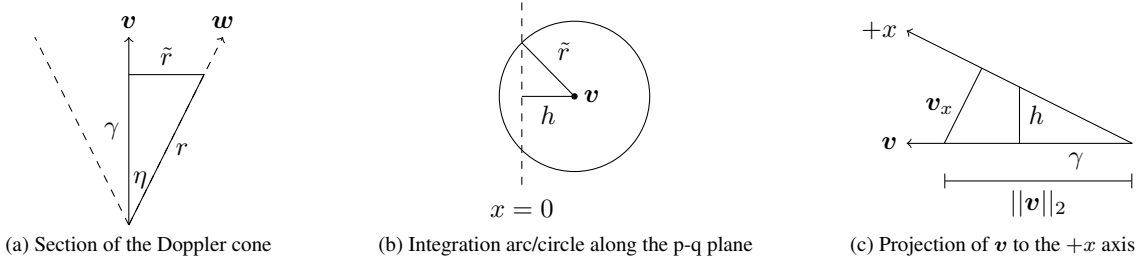
\begin{figure*}
\centering
\begin{subfigure}[b]{0.3\textwidth}
\centering
\begin{tikzpicture}
\draw[dashed,->] (0,0) -- (1.25,2.5) node[pos=1.0,above] {$\bm{w}$};
\draw[dashed] (0,0) -- (-1.25,2.5);
\draw[->] (0,0) -- (0,2.5) node[pos=0.5,left] {$\gamma$} node[pos=1.0,above] {$\bm{v}$};
\draw (0,2) -- (1,2) node[pos=0.5,above] {$\tilde{r}$} -- (0,0) node[pos=0.5,right] {$r$};
\node[] at (0.15,0.6) {$\eta$};
\end{tikzpicture}
\caption{Section of the Doppler cone}
\label{fig:geometry1}
\end{subfigure}\hspace{1em}
\begin{subfigure}[b]{0.3\textwidth}
\centering
\begin{tikzpicture}
\draw[dashed] (-0.7,1.25) -- (-0.7,-1.25) node[pos=1.0,below] {$x = 0$};
\draw (-0.7,0.714) -- (0,0) node[pos=0.5,above right] {$\tilde{r}$} -- (-0.7,0) node[pos=0.5,below] {$h$};
\draw (0,0) circle (1 cm);
\filldraw (0,0) circle (1 pt) node[right] {$\bm{v}$};
\end{tikzpicture}
\caption{Integration arc/circle along the p-q plane}
\label{fig:geometry2}
\end{subfigure}\hspace{1em}
\begin{subfigure}[b]{0.3\textwidth}
\centering
\begin{tikzpicture}
\draw[->] (0,0) -- (-3,1.5) node[pos=1.0,left] {$+x$};
\draw[->] (0,0) -- (-3,0) node[pos=1.0,left] {$\bm{v}$} node[pos=0.25,below] {$\gamma$};
\draw (-1.5,0) -- (-1.5,0.75) node[pos=0.5,right] {$h$};
\draw (-2.5,0) -- (-2.0,1.0) node[pos=0.5,left] {$\bm{v}_x$};
\draw[|-|] (-2.5,-0.5) -- (0,-0.5) node[pos=0.5,below] {$||\bm{v}||_2$};
\end{tikzpicture}
\caption{Projection of $\bm{v}$ to the $+x$ axis}
\label{fig:geometry3}
\end{subfigure}

\caption{Illustration of our integration arc equation derivation.}
\end{figure*}

\paragraph{Field of view calculation}

While radar rendering theoretically requires integrating spherically around the radar, the gain to the rear of the radar is close to zero in practice\footnote{
    Our radar uses a PCB-based antenna, which acts as a metal plane preventing the radar from seeing backwards.
}. As such, we only integrate across the front of the radar.

Our rendering equation begins by generating an orthonormal basis
\begin{equation}
\left \{ p, q, \frac{\bm{v}}{||\bm{v}||_2} \right \}
\end{equation}
for velocity vector $\bm{v}$. We select $p$ such that its direction is aligned with the positive $x$ axis (which define to be the front of the sensor), i.e. 
\begin{equation}
p \propto \begin{bmatrix}1 & 0 & 0\end{bmatrix}^T - \frac{\bm{v}_x\bm{v}}{||\bm{v}||_2^2},
\end{equation}
and define $q$ accordingly.

Each range-Doppler bin is modeled as the intersection of a sphere with radius $r$ and a cone $\langle \bm{v}, \bm{w}\rangle = d$. When intersected with the half-space corresponding with $x \geq 0$, this results in an arc or circle on the p-q plane (or the empty set if degenerate). Since we define the sensor to be facing in the $+x$ axis, and define $p$ to be aligned with $+x$, the valid integration arc will always be $(-\psi, \psi)$ for angle $\psi$ counterclockwise from the $+p$ axis, again relative to the p-q plane.

We now calculate this angle $\psi$. We begin by calculating the distance $\gamma$ to the center of the integration arc and its radius $\tilde{r}$ (Fig.~\ref{fig:geometry1}). Choosing an arbitrary direction $\bm{w}$ in our p-q-v basis, the cosine of the angle $\eta$ between $\bm{v}$ and $\bm{w}$ yields
\begin{align}
    \cos(\eta) = \frac{\gamma}{r} = \frac{\langle \bm{v}, \bm{w} \rangle}{||\bm{v}||_2||\bm{w}||_2} 
    \quad\Longrightarrow\quad 
    \gamma = \frac{rd}{||\bm{v}||_2}
\end{align}
using the definition of Doppler $d$ and orthonormality of $\bm{w}$. It follows that
\begin{align}
    \tilde{r} = \sqrt{r^2 - \frac{r^2d^2}{||\bm{v}||_2}} = r\sqrt{1 - \frac{d^2}{||\bm{v}||_2^2}}.
    \label{eq:r_tilde}
\end{align}
Next, we calculate the length of the chord corresponding to the integration arc $\pm \psi$. Let $h$ be the distance from the chord to $\bm{v}$ (Fig.~\ref{fig:geometry2}); $h$ can be specified relative to $\gamma$ and the angle $\theta$ between $\bm{v}$ and the $+x$ axis (Fig.~\ref{fig:geometry3}). It follows that
\begin{align}
    h = \gamma \tan(\theta) = \frac{rd}{||\bm{v}||_2} \left(\frac{\bm{v}_x}{\sqrt{||\bm{v}||_2^2 - \bm{v}_x^2}}\right),
\end{align}
which finally yields (Fig.~\ref{fig:geometry2})
\begin{align}
    \psi &= \arccos\left(\frac{h}{\tilde{r}}\right).
    \label{eq:psi}
\end{align}
We also must account for three degenerate cases:
\begin{itemize}
    \item When $d > ||\bm{v}||_2$, the Doppler bin exceeds the velocity (and Eq.~\ref{eq:r_tilde} becomes imaginary).
    \item When $h > \tilde{r}$, the entire doppler cone is pointing behind the radar (i.e. the $x>0$ half-space no longer intersects with the cone; Eq.~\ref{eq:psi} is undefined).
    \item When $h < -\tilde{r}$, the entire doppler cone is in front of the radar. In this case, we instead have $\psi = \pi$.
\end{itemize}

\paragraph{Volume of integration region}

\begin{figure}
\centering
\begin{tikzpicture}
\draw[->] (0,0) -- (0,2.5) node[pos=1.0,above] {$\bm{v}$};
\draw[->] (0,0) -- (1.25,2.165) node[pos=1.0,above right] {$d_j + \Delta D$};
\draw[->] (0,0) -- (2.165,1.25) node[pos=1.0,right] {$d_j$} node[pos=0.6,below right] {$r_i$} node[pos=0.8,below right] {$r_i + \Delta R$};
\draw (1.732,1.0) arc (30:60:2);
\draw (1.386,0.8) arc (30:60:1.6);
\draw[dashed] (0,1.386) -- (0.8,1.386) node[pos=0.5,above] {$\tilde{r}$};
\node[] at (1.27,1.27) {$A$};
\draw[->] (0,0.9) arc (90:50:0.9);
\node[] at (0.62,0.56) {$\eta$};
\end{tikzpicture}
\caption{Annular section $A$ for the revolved solid approximation of Eq.~\ref{eq:doppler-integration}. $\eta$ is the angle to the velocity vector $\bm{v}$.}
\label{fig:geometry4}
\end{figure}
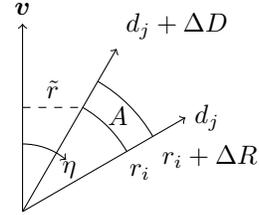

Expanding on Eq.~\ref{eq:rendering-integral} more explicitly using our integration bound $\psi$, the radar return $\bm{Y}(r_i, d_j, k)$ can be expressed as the circular integral
\begin{align}
\bm{Y}(r_i, d_j, k) \propto \tilde{r}\int\displaylimits_{-\psi \leq \phi \leq \psi}
C(r, k, \bm{w}(\phi)) \ d\phi
\label{eq:circularint}
\end{align}
for projection $\bm{w}$ to the integration arc.

While our rendering equation (Eq.~\ref{eq:rendering-integral}) makes a ``thin-shell'' assumption, we must also examine the effect of \textit{bin size} in order to correctly determine which constants to apply to our approximation. Let $\Delta R$ and $\Delta D$ be the width of each range and Doppler bin. Instead of using a ``point'' approximation of $C$, we instead derive the integral based on the revolved solid of an annular section $A$ (Fig.~\ref{fig:geometry4}). Using again the fact that $\cos(\eta) = d/||\bm{v}||_2$, we arrive at the approximation
\begin{align}
    dA &= r_i \ dr d\theta
    = \frac{r_i}{||\bm{v}||_2 \sqrt{1 - \frac{d_j^2}{||\bm{v}||_2^2}}} \ dr dd \nonumber \\
    &= \frac{r_i^2}{||\bm{v}||_2\tilde{r}} \ dr dd,
\end{align}
which yields the net correction of $r_i^2/||\bm{v}||_2$ found in Eq.~\ref{eq:doppler-integration} when combined with Eq.~\ref{eq:circularint}.

\paragraph{Array factor and antenna gain}

In our rendering process, we begin by applying the antenna gain for the radar as a whole, which is specified by the AWR1843Boost datasheet \cite{ti:AWR1843AOP}. After performing the azimuth FFT, each azimuth bin (i.e. transmit-receive pair) then acts as an independent beamformed radar with a narrower azimuth gain; as such, we also apply the array factor to each azimuth bin \cite{richards2010principles}.

\subsection{Optimized Radar Rendering}
\label{appendix:rendering}

\begin{figure}
    \centering
    \includegraphics[width=\columnwidth]{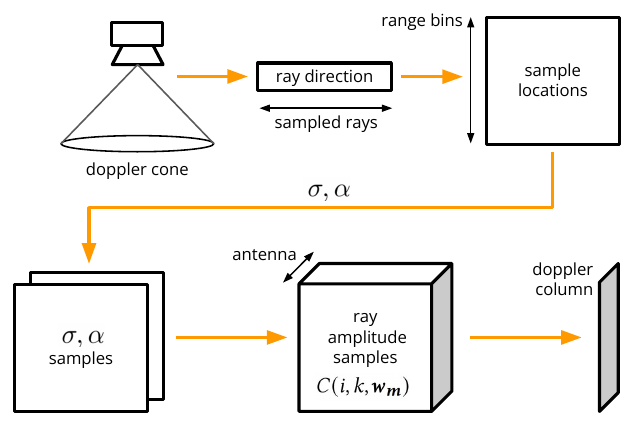}
    \vspace{-2em}
    \caption{Radar rendering dataflow.}
    \label{fig:computation}
\end{figure}

Figure~\ref{fig:computation} shows our optimized rendering dataflow:
\begin{enumerate}[noitemsep,topsep=0pt,leftmargin=*]
    \item We use each Doppler ``column'' across all antennas as a sample which has (range, antenna) dimensions.
    \item For this Doppler, we sample a fixed (128) number of rays (i.e. from $(-\psi, \psi)$, projecting to 3D space via the v-p-q basis); we trace each ray (Eq.~\ref{eq:raytracing}), sampling $\sigma$ and $\alpha$ once at each range bin for a total of (Doppler, ray samples) times.
    \item For each antenna, we apply the corresponding gain and sum across the Doppler samples (Eq.~\ref{eq:doppler-integration}), yielding (range, antenna) outputs for this Doppler column.
\end{enumerate}
In total, we sample the field  (range, Doppler, ray samples) times per image per epoch, which is comparable to the (width, height, range samples) times in NeRF. With the dimensions of images in our datasets (128 range bins, 256 Doppler bins, 128 ray samples, and 8 antenna), this equates to only 16 field samples per output value --- less than most visual NeRF methods today such as Nerfacto \cite{tancik2023nerfstudio}, which samples 48 points per pixel.

\subsection{Training and other Hyperparameters}
\label{appendix:training}

\paragraph{Implementation and Training}

DART is implemented using JAX \cite{jax2018github}, which we tune for training on the RTX 4090 (24GB of VRAM). In our implementation, we perform the ray-tracing cumulative product in the log-space as an exp-sum-log, learning the log-transmittance instead. We also absorb any constants (including the square for transmittance) into $\sigma$ and $\alpha$. 

We trained our $(\sigma, \alpha)$ field function using stochastic gradient descent with the Adam optimizer and an $l_1$ objective with a learning rate of 0.01 for 3 epochs with a batch size of 1024 ``Doppler columns'' (i.e. all range-Doppler-antenna bins with the same Doppler value in a given image; see Sec.~\ref{appendix:rendering}). For each Doppler column, we sampled 128 rays along the arc defined by that Doppler bin intersected with the half-sphere corresponding to the front of the radar (since our radar is forward-facing).

\paragraph{Architecture Hyperparameters}

For our base model, we use the Instant Neural Graphics Primitive \cite{muller2022instant} with the following parameters:
\begin{itemize}
    \item Hash table size: $2^{20}$.
    \item Number of features per hash table level: 2.
    \item Hash table resolution scale factor: $2^{0.43} \approx 1.347$ (selected to avoid ``aliasing'' from different hash table levels lining up to multiples of each other).
    \item Number of hash table levels: 12.
    \item Grid resolution: 25cm (coarsest) to 0.94cm (finest).
    \item Output MLP: 2 hidden layers of 64 and 32 units.
\end{itemize}

\section{Datasets and Evaluation}
\label{appendix:evaluation}

In this section, we provide additional details about our data collection system (\ref{appendix:system}), collected radar traces (\ref{appendix:datasets}), baselines (\ref{appendix:baselines}), and evaluation methodology (\ref{appendix:metrics}).

\begin{table*}
\small
\centering
\vspace{1em}
\begin{tabular}{c c c p{1.25\columnwidth}}
    \toprule
    Dataset & Length & Training Time & Description \\
    \toprule
    Lab 1 & 4:55 & 8:43 & A lab space with 5 boxes of varying material placed inside. \\
    Lab 2 & 4:24 & 7:22 & \\
    \hline
    Office 1 & 10:37 & 15:14 & An office space with cubicles and an enclosed meeting room. \\
    Office 2 & 15:39 & 22:18 & \\
    \hline
    Yard & 7:38 & 12:13 & A backyard and adjacent detached garage. \\
    \hline
    House 1 & 7:55 & 9:53 & \multirow{2}{1.25\columnwidth}{A early 20th century house with two above-ground floors. Traces include only the ground floor, and both the ground and 2nd floor.}\\
    House 2 & 9:48 & 12:54 &\\
    \hline
    Apartment 1 & 6:04 & 9:26 & \multirow{2}{1.25\columnwidth}{A late 20th century high rise one-bedroom apartment; traces include just the open-plan living area, and both the living area and bedroom.}\\
    Apartment 2 & 6:38 & 10:30 &\\
    \hline
    Rowhouse 1 & 7:41 & 10:48 & \multirow{2}{1.25\columnwidth}{Traces of varying lengths and room inclusions in a 2-bedroom apartment with an open-plan living space in a low-rise townhouse.} \\
    Rowhouse 2 & 7:37 & 10:41 &\\
    Rowhouse 3 & 10:51 & 15:17 &\\
    \toprule
\end{tabular}
\vspace{-1em}
\caption{Total trace length and training times (minutes) for our method on each trace; the training time of our method is between 1-2$\times$ the data collection time. While we captured multiple traces from each location, each trace was collected independently.}
\label{tab:datasets}
\end{table*}

\subsection{Data Collection System}
\label{appendix:system}

\begin{figure}
    \centering
    \includegraphics[width=\columnwidth]{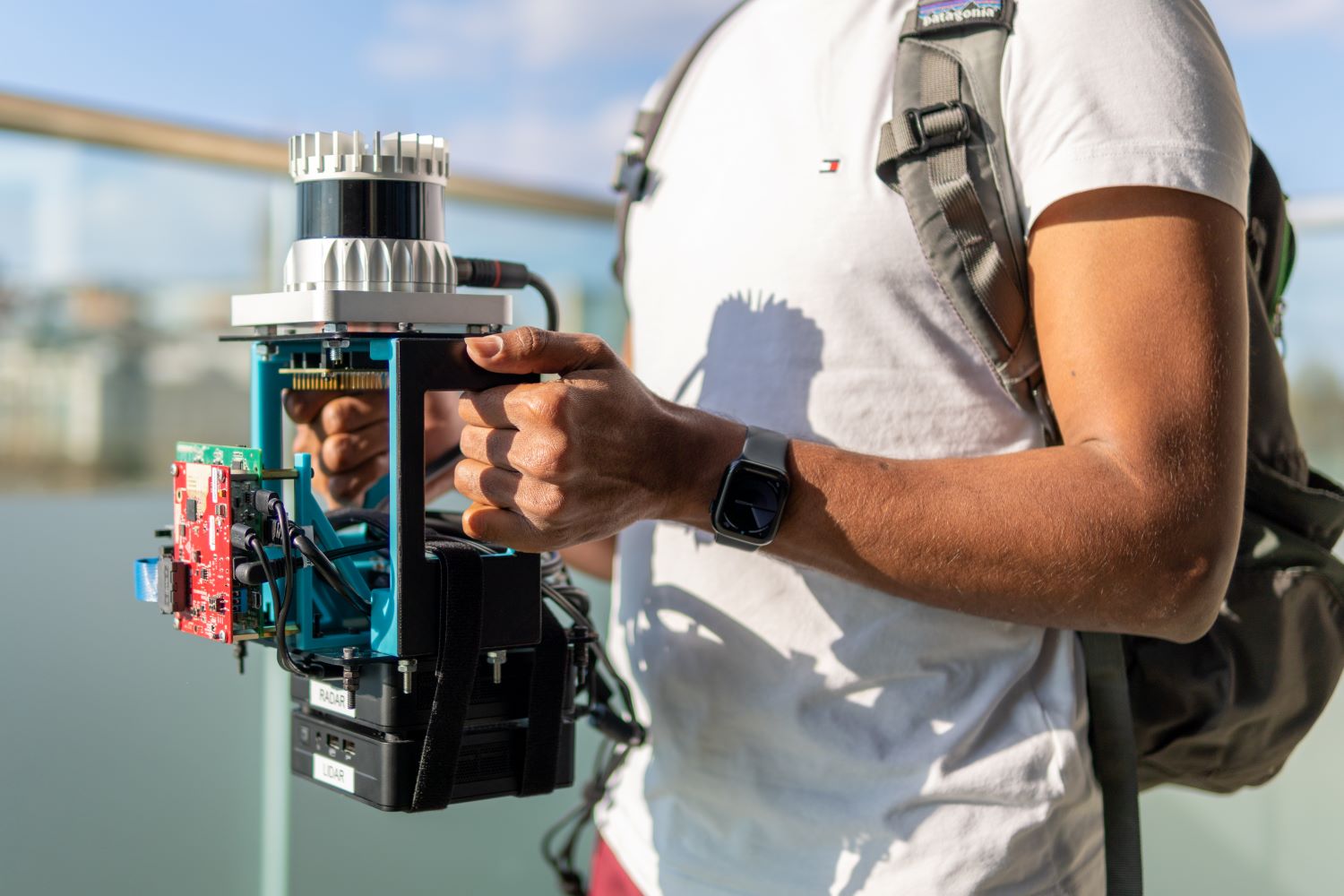}
    \vspace{-0.5em}
    \caption{Our data collection system consists of a handheld rig (Fig.~\ref{fig:data-collection-rig}) and a backpack which carries power equipment.}
    \label{fig:backpack}
\end{figure}

\begin{figure}
    \centering
    \includegraphics[width=\columnwidth]{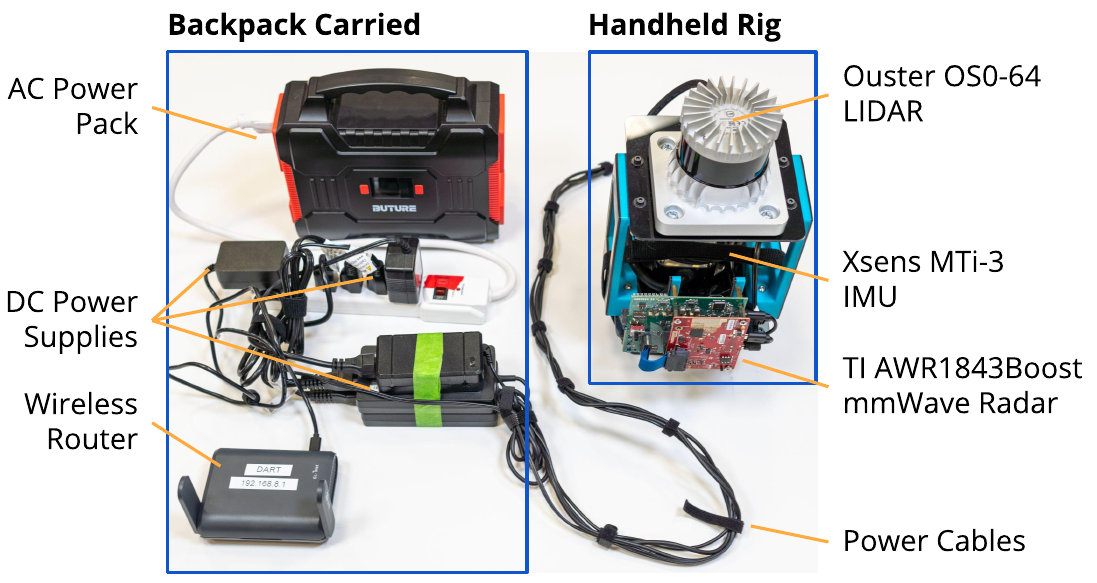}
    \vspace{-2em}
    \caption{Components of our data collection system.}
    \label{fig:equipment}
\end{figure}

Our data collection rig (Fig.~\ref{fig:backpack}) is handheld and fully portable, and consists of a TI AWR1843Boost mmWave radar with a DCA1000EVM capture card for recording raw I/Q frames. The data collection rig also includes an Ouster OS0-64 64-beam LIDAR and a Xsens MTi-3 IMU, which are used for our LIDAR-based radar simulator baseline (Appendix~\ref{appendix:baselines}) as well as pose estimates using Cartographer SLAM \cite{hess2016real}. These components, along with the backpack-carried power and control components, are shown in Fig.~\ref{fig:equipment}.

\paragraph{Radar Configuration}

We configured the radar for inter-chirp period of 1ms with 512 samples per chirp, and processed the data into range-Doppler images with a frame length of 256 chirps computed on a rolling window with a stride of 64 chirps. This gives a maximum Doppler of $\pm 0.95$ m/s across 256 Doppler bins and a maximum range of 21.6m across 512 range bins, which we truncated to the first 128 range bins (5.4m) after removing bins with negative range after radar calibration.

\paragraph{Radar Processing}

To avoid degenerate Doppler bins at low speeds (violating our thinness assumption) and aliased speeds exceeding the maximum Doppler velocity, we removed frames with speeds below 0.2m/s and above 0.95m/s from our training and validation datasets. We also use all 8 azimuth antennas by configuring transmitters to operate in a time-division multiplex mode.

\paragraph{Pose Processing}

DART relies on accurate velocity estimates; however, SLAM systems such as Cartographer are typically not designed to produce accurate or continuous velocity estimates, especially at the frequency that we generate range-Doppler images (64ms). This can lead to erratic velocity estimates, especially if loop closure results in discontinuous pose estimates.

To calibrate the estimated speed, we implemented a speed estimation algorithm which uses simple thresholding to detect the maximum velocity detected in a range-Doppler image. Assuming a static scene, this then indicates the speed of the radar. Using this estimate, we tuned a gaussian smoothing parameter applied to the SLAM pose estimates to produce our final velocity estimates.

In our datasets, we observed that false or sudden relocalizations can also cause erratic speed estimates. Thus, as a final check, we calculated the acceleration of the final speed estimates, and excluded frames (and $\pm15$ neighbors) if the acceleration exceeded $2.0 \ \text{m}/\text{s}^2$. Datasets that we collected that included a high frequency of likely SLAM system failures were then either recollected or excluded.

\subsection{Collected Traces}
\label{appendix:datasets}

\begin{figure*}
\centering
\begin{subfigure}[b]{0.45\textwidth}
    \centering
    \includegraphics[height=1.5in]{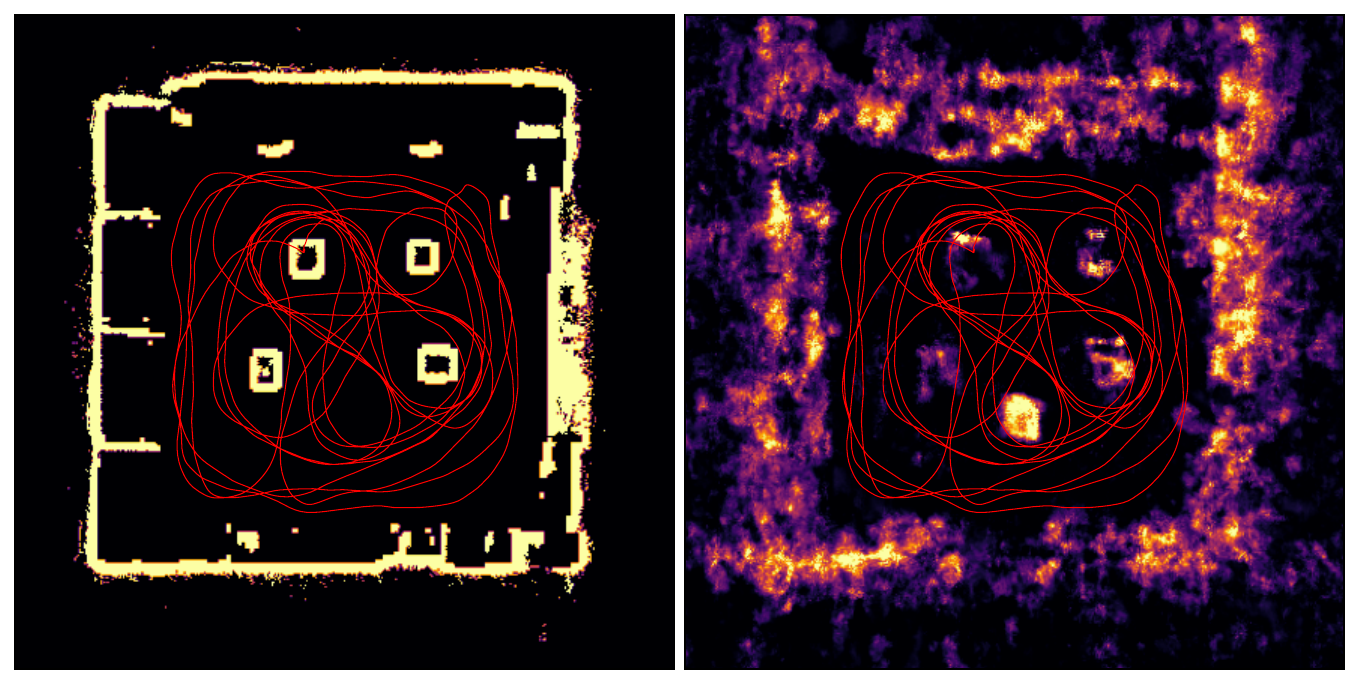}
    \caption{Lab 1, 2}
\end{subfigure}
\begin{subfigure}[b]{0.38\textwidth}
    \centering
    \includegraphics[height=1.5in]{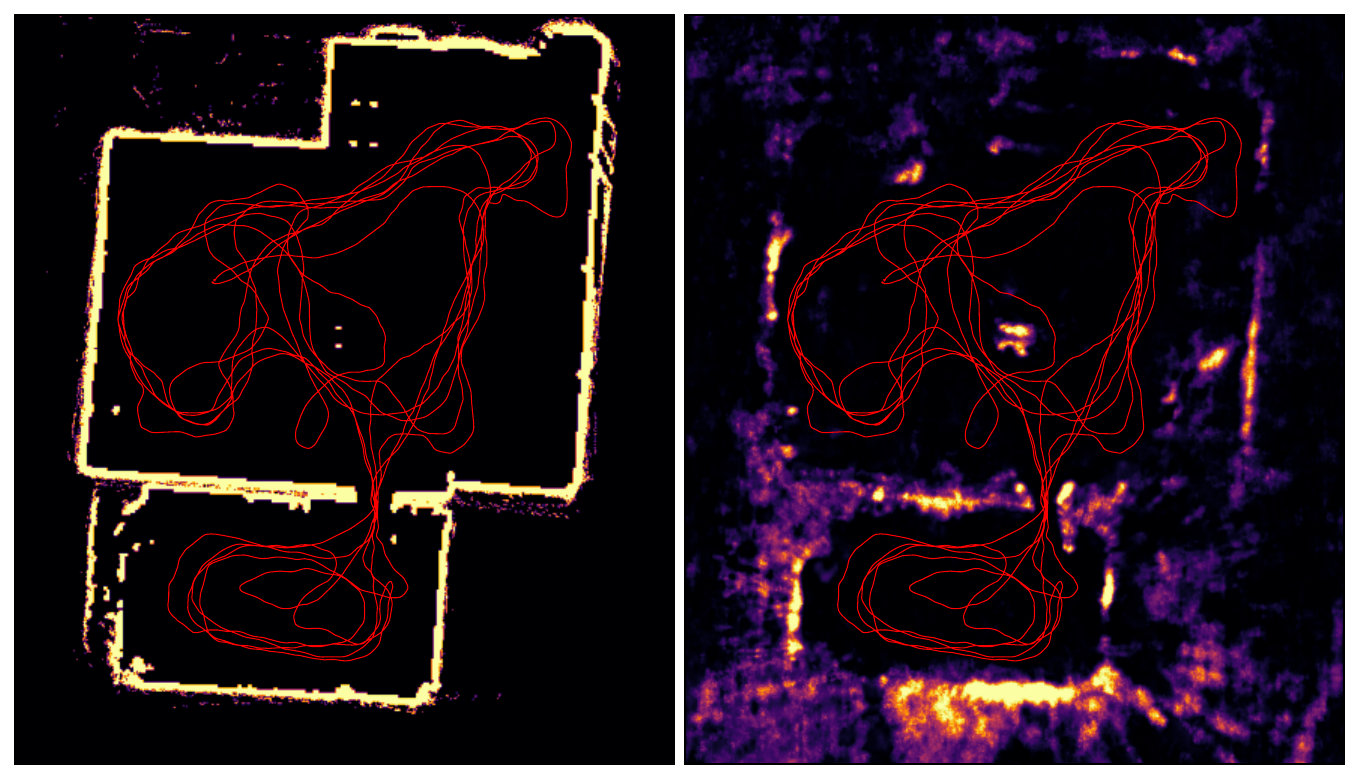}
    \caption{Yard}
\end{subfigure}

\begin{subfigure}[b]{0.55\textwidth}
    \centering
    \includegraphics[height=1.5in]{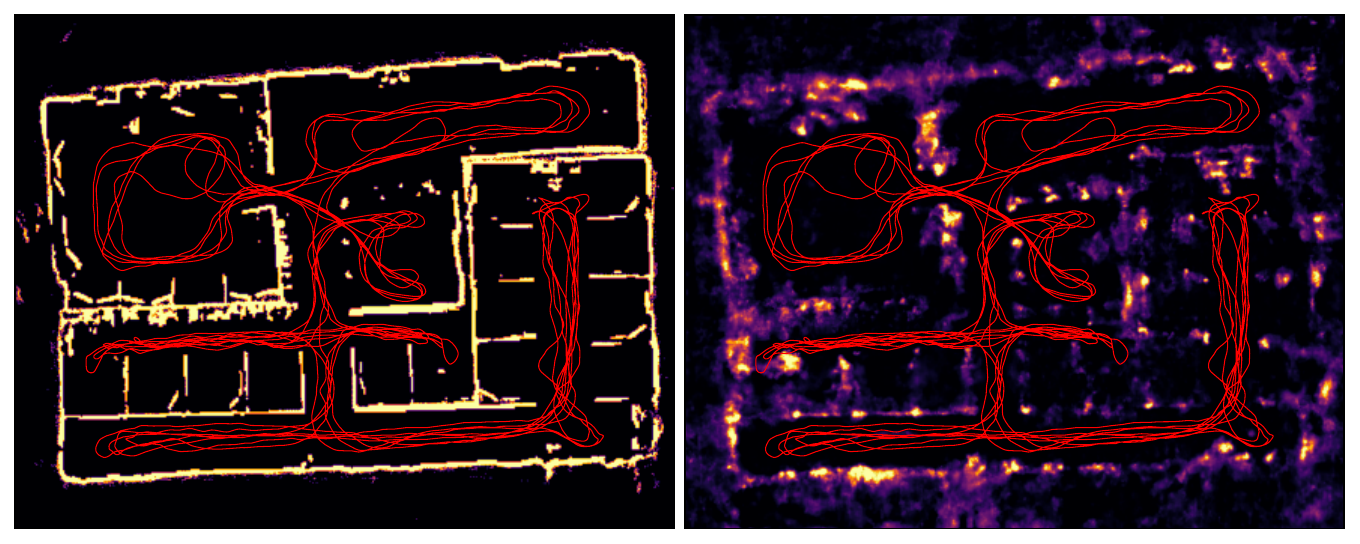}
    \caption{Office 1, 2}
\end{subfigure}
\begin{subfigure}[b]{0.34\textwidth}
    \centering
    \includegraphics[height=1.5in]{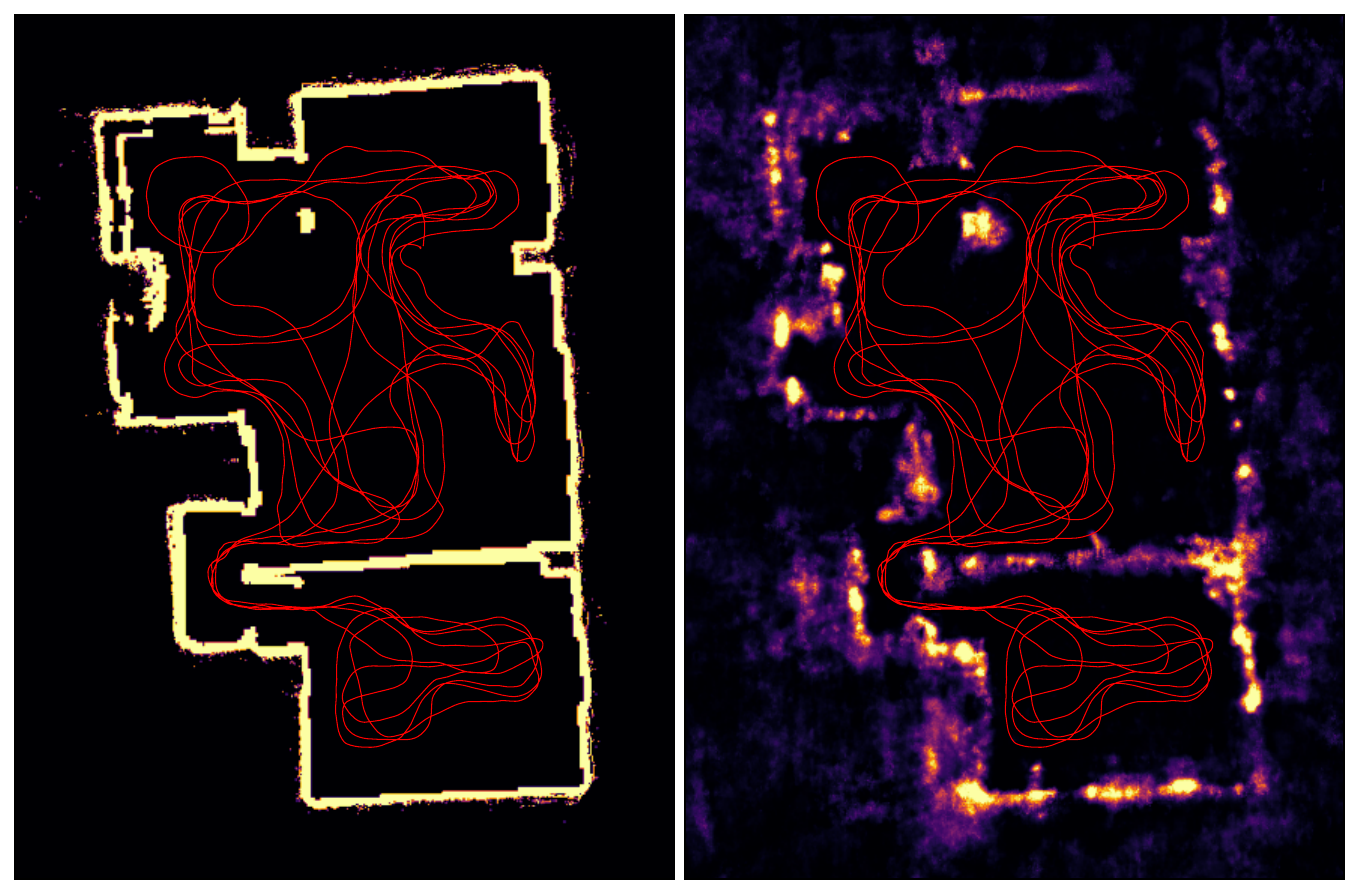}
    \caption{Apartment 1, 2}
\end{subfigure}

\begin{subfigure}[b]{0.49\textwidth}
    \centering
    \includegraphics[height=1.5in]{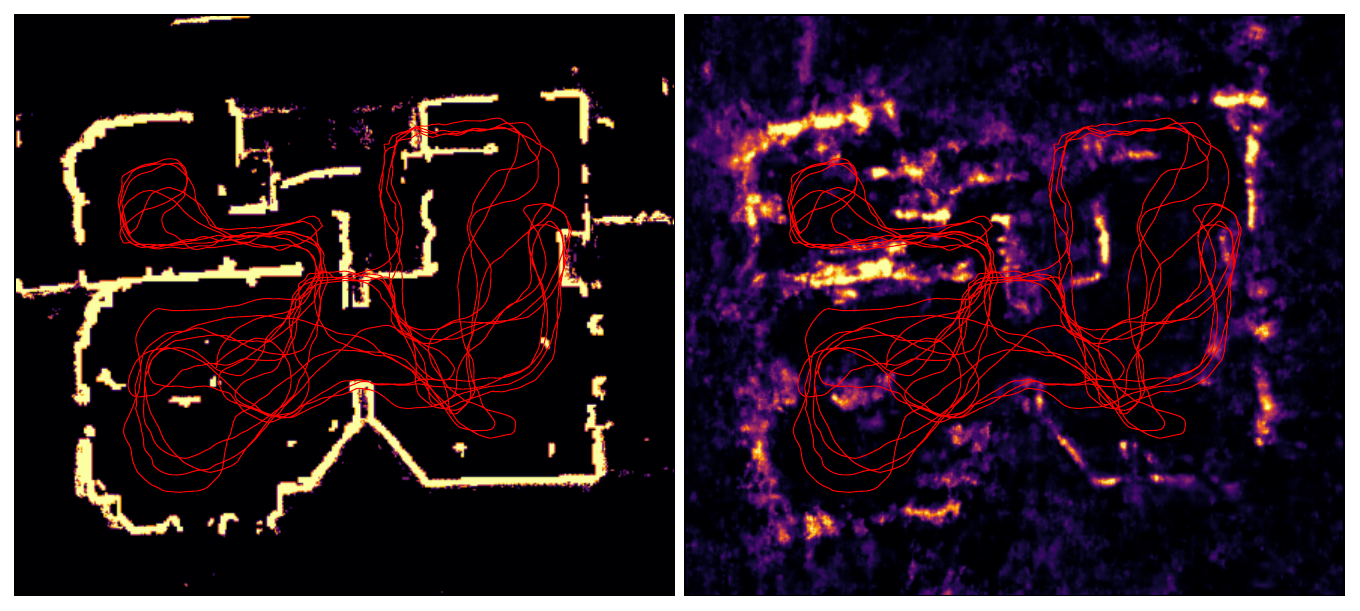}
    \caption{House 1, 2 (1st floor)}
\end{subfigure}
\begin{subfigure}[b]{0.49\textwidth}
    \centering
    \includegraphics[height=1.5in]{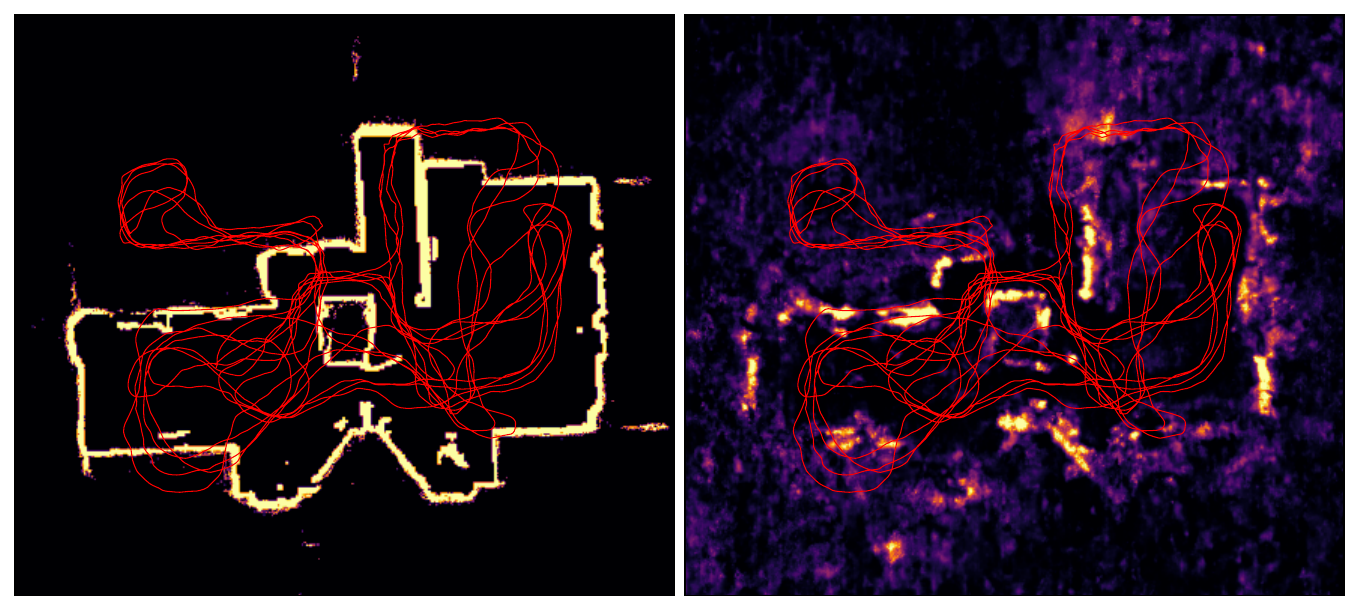}
    \caption{House 1, 2 (2nd floor)}
\end{subfigure}

\begin{subfigure}[b]{0.8\textwidth}
    \centering
    \includegraphics[height=1.5in]{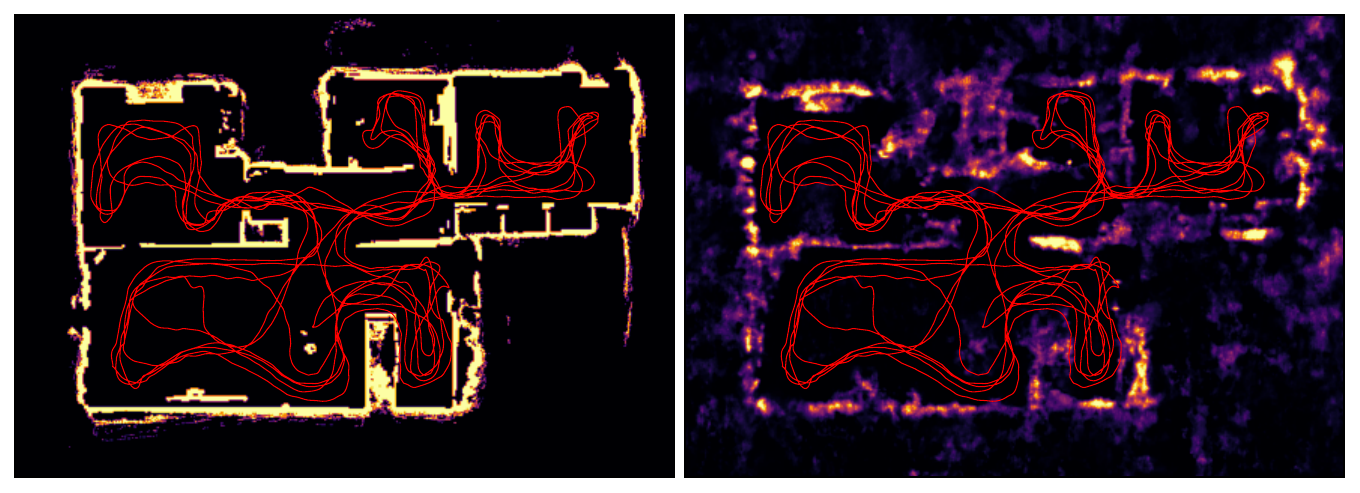}
    \caption{Rowhouse 1, 2, 3}
\end{subfigure}
\vspace{-0.5em}
\caption{Lidar maps (left), radar tomography maps (right), and sample trajectories (red) on the 6 locations used to collect our novel radar view synthesis dataset. Multiple traces were collected from each location, with each trace being completely independent (and in some cases visiting different rooms). Note that \textit{House} contains two floors; the trajectory shown here is not separated by floor.}
\label{fig:maps}
\end{figure*}

Our 12 collected radar novel view synthesis traces include 6 unique environments across residential and commercial spaces of various construction. Each trace consists of several passes through each scene with different paths, velocities, and orientations, and is collected with a full pass through the scene at the end to enable a convenient test set. We summarize these traces in Table~\ref{tab:datasets}, and provide a visual map of the lidar scan and \name's radar tomography in Fig.~\ref{fig:maps}.

\paragraph{Trajectory} Like many off-the-shelf radars, the TI AWR1843 radar is designed for automotive applications, and has a relatively narrow vertical field of view ($\pm$20\textdegree, compared to $\pm$50\textdegree \ horizontally). As such, we took multiple passes with varying attitude (e.g. horizontal, up, and down) in order to ensure full 3D coverage.

\paragraph{Velocity} We found Cartographer to be most stable for our handheld data collection platform when moving smoothly at $\approx$0.5m/s, and targeted this speed during each trace. This also corresponds to the velocity range (0.2m/s -- 0.95m/s) which we tuned our radar system for.

\subsection{Baseline Details}
\label{appendix:baselines}

In this section, we provide additional details about the implementation of our baselines.

\paragraph{Lidar Scan-based Simulator}

Using Cartographer \cite{hess2016real}, we created an occupancy grid of each scene with a 2cm resolution. Each occupied cell was assigned a reflectance of 1 unit, and a transmittance of 0 (i.e. fully opaque). We then used a ray-tracing simulator (based on our rendering equation) to simulate radar images, with trilinear interpolation applied to the grid during sampling. This simulator used the same settings as DART (128 rays per Doppler bin, 128 range bins, simultaneous range-antenna rendering).

\paragraph{CFAR Point Cloud Aggregation}

We used the CFAR implementation provided by the Matlab Phased Array System Toolbox \cite{cfarmatlab} with a false alarm rate of $0.01$ to select discrete reflectors. Each reflector was assigned an estimated azimuth angle using direction of arrival estimation, projected to 3D space, and aggregated into a common point cloud along with their estimated reflectance amplitude.

Since finite element modeling is impractical with tens of thousands of CFAR points, we projected each reflector to a 5cm grid (which must be coarser than the 2cm lidar-based simulator grid due to lower point cloud density), taking the maximum gain for each point when multiple fall in the same grid cell. Finally, we used the same simulation settings as our lidar baseline to generate range-Doppler images. Note that since CFAR has no way of measuring the transmittance of points in space, we set all grid cells to be fully transparent (i.e. transmittance $\alpha = 1$).

For our comparison of the CFAR point cloud with \name's sampled tomography images (Fig.~\ref{fig:tomography}), we qualitatively tuned the visualization to show sharp outlines and the general geometry of the scene to the best of our ability:
\begin{itemize}
    \item The scatter plot shows points within $\pm 10$cm of the reference plane to ensure we capture a sufficient density of strong reflectors.
    \item To reduce clutter due to spurious points in the CFAR point cloud, we then removed the 80\% weakest reflectors, and scaled the color range in our visualization so that points close to the 80\% cutoff point have a color similar to the plot background.
\end{itemize} 

\subsection{Evaluation Metrics}
\label{appendix:metrics}

In order to evaluate DART and our baselines, we calculate the SSIM of each valid\footnote{
    Since our Doppler FFT ``wraps'' once we exceed our maximum doppler velocity, we must exclude these frames. Our method also does not work on stationary radars (as we note in our limitations), so frames with a velocity near zero are also excluded.
} range-Doppler-antenna frame (treating each antenna as a different channel) after normalizing each image to a set [0, 1] range. Then, to calculate error bars and evaluate the statistical significance of our results, we estimate the effective sample size of each trace in order to calculate standard errors and perform a paired z-test.

\paragraph{Image Scaling} Not all of our baselines are capable of producing accurately scaled output images. Since SSIM requires the input dynamic range to be properly scaled, we scale all images relative to the ground truth prior to SSIM calculation as follows:
\begin{enumerate}
    \item We clip extreme values from the ground truth images ($0.1\%$ and $99.9\%$ percentiles), and normalize this range to $[0, 1]$, which we use as the dynamic range for SSIM calculation.
    \item We compute the $l_2$ ``optimal scaling'' multiplier between the predicted $\hat{y}$ and actual $y^*$ images using scale factor
    \begin{align}
        \arg\min_\xi ||y^* - \xi\hat{y}||_2^2 = \frac{\hat{y}^Ty^*}{\hat{y}^T\hat{y}}
    \end{align}
    and apply this to $\hat{y}$. We then clip $\xi\hat{y}$ to the same thresholds as $y^*$, and normalize to [0, 1].
    \item We then compute the SSIM, using a dynamic range of 1.
\end{enumerate}

\paragraph{SSIM Calculation} Range-Doppler-antenna radar images are mostly sparse, and even contain large regions where nonzero values cannot theoretically be observed (i.e. when Doppler velocities exceed the current actual speed for a sensor in a stationary scene). As such, a naive application of SSIM leads to misleadingly high SSIM, since the SSIM between empty regions is exactly 1.0.

To correct for this, we exclude empty regions of our image, which are defined as areas where the pixel sample mean is less than a set threshold ($\varepsilon = 0.005$ in our dataset).

\begin{figure*}
\centering
\includegraphics[width=\textwidth]{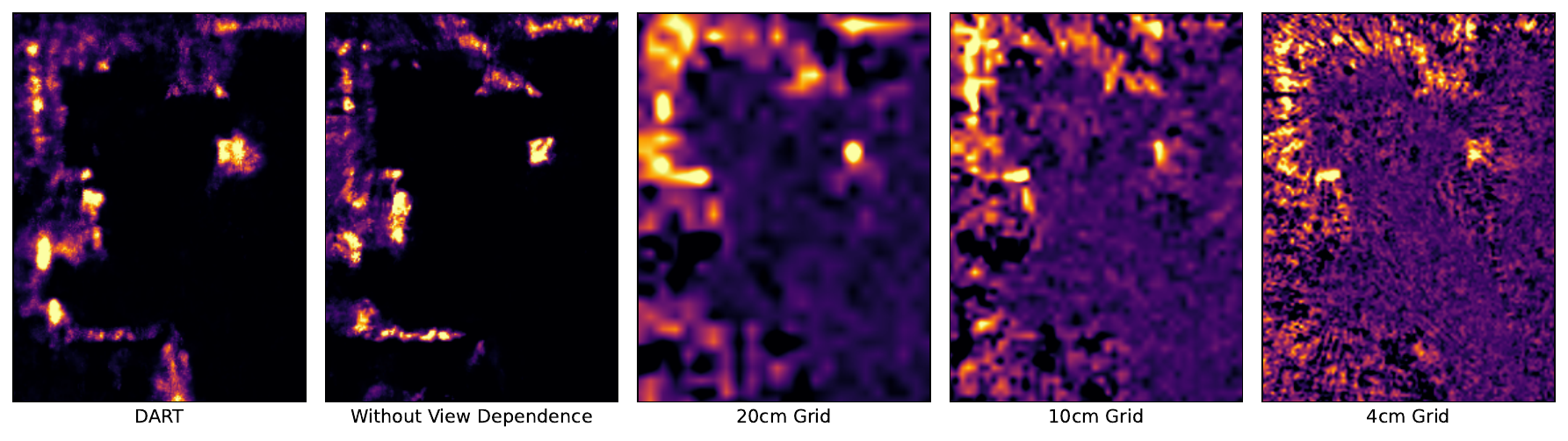}
\vspace{-2.0em}
\caption{Sample tomographic maps of the reflectance ($\sigma$) compared to using a simple grid with varying resolutions. Coarse grids result in blurry tomographic images, while finer grids result in ``holes'' in walls and other surfaces. By using an adaptive grid, \name\ is able to learn tomographic maps which are reasonably sharp but also largely continuous.}
\label{fig:ablation_grid}
\end{figure*}

\begin{figure*}
\centering
\includegraphics[width=\textwidth]{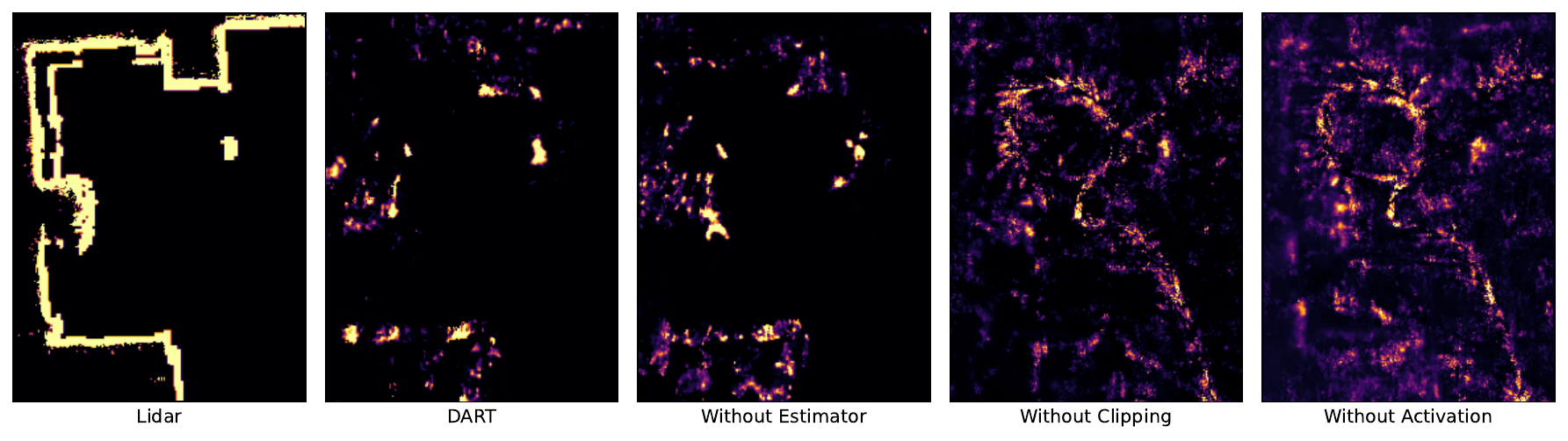}
\vspace{-2em}
\caption{Sample tomographic maps of the transmittance ($\alpha$) learned by \name\ (compared to a Lidar occupancy grid), along with ablations on our representation for $\alpha$ without the gradient estimator (\textit{Without Estimator}), without clipping $\alpha$ when $\sigma < 0.05$ (\textit{Without Clipping}), and without any activation treatment (\textit{Without Activation}).}
\label{fig:ablation_clip}
\end{figure*}

\paragraph{Sample Size Correction} The standard error of the mean (SE) of a measurement $X$ is defined as $\text{std}(X)/\sqrt{N}$. However, this assumes each of the $n$ samples are independent and identically distributed.

Intuitively, sampling the same signal with greater frequency does not give any additional insight into the general performance of a method. Thus, in time series signals such as ours (i.e. SSIM over time), the number of samples must be \textit{corrected} to estimate the effective sample size, which we substitute for $N$. This is given by \cite{robert1999monte}
\begin{equation}
    N_{\text{eff}} = \frac{N}{1 + 2\sum_{t=1}^\infty \rho_t}
\end{equation}
for autocorrelation $\rho_t$ (where $t$ is the delay). In our calculations, we estimate the sum up to $t = N/2$, and clip $\rho_t$ estimates to positive values\footnote{Negative $\rho_t$ can occur for large delays $t$ due to broader trends in our dataset, e.g. scanning different rooms or directions}.

\paragraph{Paired z-Test} Since each method is evaluated on exactly the same test frames, we use a paired z-test to evaluate the statistical significance of our results. Specifically, we provide error bars for a one-sided p-value of 0.05 (i.e. indicating 95\% confidence that one method is better than another).

\section{Additional Results}

\subsection{Ablations}
\label{appendix:ablations}

In addition to view dependence, we ran additional ablations showing the impact of \name's implicit representation and transmittance activation function.

\paragraph{Grid Size and Adaptive Grid} To demonstrate the benefit of an (implicit) adaptive grid, we trained a version of \name\ using a fixed grid of 20cm, 10cm, and 4cm resolutions. \name's adaptive grid leads to more accurate radar simulations compared to fixed grids (Table~\ref{tab:ablation_grid}). Examining tomographic slices of the learned reflectance reveals the reason: a coarse grid leads to a blurry image (and therefore blurry synthesized radar range-Doppler images), while a fine grid leads to holes in the learned reflectance (Fig.~\ref{fig:ablation_grid}).

\begin{table}
\small
\centering
\begin{tabular}{l l l}
\toprule
Method & Mean SSIM & Relative SSIM \\
\toprule
\textbf{\name} & 0.636 $\pm$ 0.012 & --- \\
No View Dep. & 0.614 $\pm$ 0.015 & 0.022 $\pm$ 0.005 \\
20cm Grid & 0.591 $\pm$ 0.015 & 0.046 $\pm$ 0.004 \\
10cm Grid & 0.580 $\pm$ 0.017 & 0.056 $\pm$ 0.006 \\
4cm Grid & 0.564 $\pm$ 0.019 & 0.073 $\pm$ 0.008 \\
\toprule
\end{tabular}
\vspace{-0.5em}
\caption{Mean and relative SSIM (to \name) of different grid sizes; we include a baseline which does not use view dependence for comparison since a simple reflectance and transmittance grid does not represent view dependence.}
\vspace{-0.5em}
\label{tab:ablation_grid}
\end{table}

\paragraph{Transmittance Activation Function}

While we did not run full ablations for each element of the design of \name's activation function (Sec.~\ref{appendix:alpha}), we trained ablations on an example scene (Fig.~\ref{fig:ablation_clip}) for the activation function as a whole in addition to ablations on our transmittance clipping procedure and gradient estimator.

\begin{table*}[t]
\small
\centering
\begin{tabular}{c c c c c}
\toprule
Dataset & DART & Lidar & Nearest & CFAR \\
\toprule
Lab 1 & 0.676 & 0.438 (0.238 $\pm$ 0.017) & 0.490 (0.186 $\pm$ 0.033) & 0.571 (0.105 $\pm$ 0.013) \\
Lab 2 & 0.680 & 0.471 (0.209 $\pm$ 0.012) & 0.468 (0.212 $\pm$ 0.045) & 0.566 (0.114 $\pm$ 0.014) \\
Office 1 & 0.642 & 0.461 (0.181 $\pm$ 0.014) & 0.482 (0.160 $\pm$ 0.026) & 0.555 (0.087 $\pm$ 0.013) \\
Office 2 & 0.669 & 0.478 (0.190 $\pm$ 0.025) & 0.449 (0.220 $\pm$ 0.028) & 0.562 (0.107 $\pm$ 0.019) \\
Rowhouse 1 & 0.622 & 0.458 (0.164 $\pm$ 0.023) & 0.471 (0.151 $\pm$ 0.031) & 0.535 (0.087 $\pm$ 0.018) \\
Rowhouse 2 & 0.619 & 0.458 (0.161 $\pm$ 0.018) & 0.450 (0.169 $\pm$ 0.036) & 0.529 (0.090 $\pm$ 0.025) \\
Rowhouse 3 & 0.620 & 0.462 (0.158 $\pm$ 0.017) & 0.480 (0.140 $\pm$ 0.026) & 0.540 (0.080 $\pm$ 0.022) \\
House 1 & 0.626 & 0.471 (0.155 $\pm$ 0.017) & 0.462 (0.164 $\pm$ 0.025) & 0.549 (0.077 $\pm$ 0.013) \\
House 2 & 0.629 & 0.474 (0.155 $\pm$ 0.016) & 0.459 (0.170 $\pm$ 0.026) & 0.524 (0.105 $\pm$ 0.016) \\
Yard & 0.615 & 0.457 (0.158 $\pm$ 0.020) & 0.466 (0.149 $\pm$ 0.033) & 0.538 (0.077 $\pm$ 0.022) \\
Apartment 1 & 0.620 & 0.464 (0.155 $\pm$ 0.015) & 0.459 (0.160 $\pm$ 0.032) & 0.533 (0.087 $\pm$ 0.015) \\
Apartment 2 & 0.618 & 0.460 (0.158 $\pm$ 0.022) & 0.479 (0.139 $\pm$ 0.030) & 0.537 (0.081 $\pm$ 0.018) \\
\textbf{Overall} & \textbf{0.636} & \textbf{0.463 (0.174 $\pm$ 0.013)} & \textbf{0.468 (0.168 $\pm$ 0.012)} & \textbf{0.545 (0.091 $\pm$ 0.006)} \\
\toprule
\end{tabular}
\vspace{-0.5em}
\caption{The mean SSIM of each dataset for each method along with the improvement of \name\ over each baseline and 95\% (one-sided) confidence intervals calculated using a z-Test with a effective sample size-adjusted standard error. CFAR is the next best baseline in all cases, followed by \textit{Lidar} and \textit{Nearest} which perform similarly depending on the similarlity between the train and test sets.}
\label{tab:ssim_full}
\end{table*}

\begin{figure*}[t]
\includegraphics[width=\textwidth]{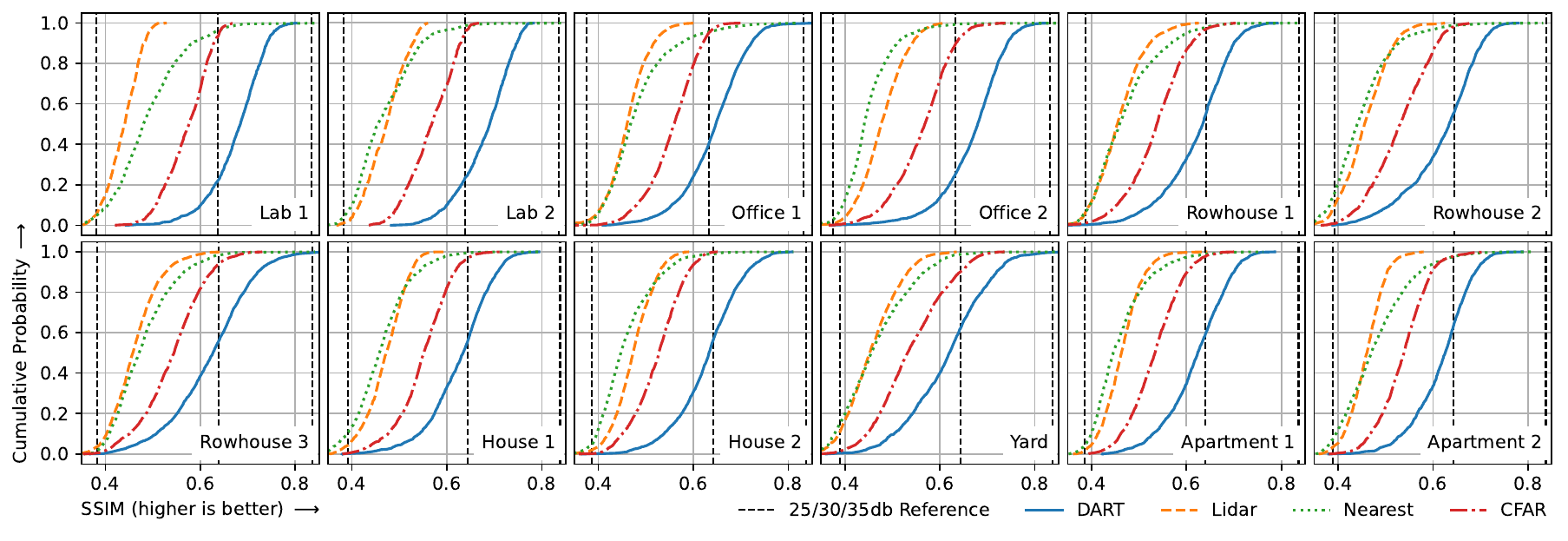}
\vspace{-2.5em}
\caption{SSIM CDFs for each of the 12 traces (down and to the right is better), along with the mean SSIM of Gaussian noise with 25/30/35db PSNR for reference. DART achieves much higher SSIMs than each baseline on all traces in our dataset, which is roughly equivalent to gaussian noise with a PSNR of 30db.}
\vspace{-0.5em}
\label{fig:ssim_cdf}
\end{figure*}

\begin{itemize}
     \item Without any activation function on $\alpha$, \name\ can greatly overfit the training data (e.g. with $\alpha > 1$). This can be seen in a very ``cloudy'' transmittance map, along with a ``trail'' of low transmittance that appears to follow the training trajectory. These artifacts are similar to ``foggy'' artifacts that can be seen in many NeRF techniques.
     \item Adding the activation function $\alpha = \exp(\min(0, \alpha'))$ with our gradient estimator results in a cleaner map; however, the trail along the training trajectory still remains, since this only addresses some forms of overfitting.
     \item To prevent \name\ from overfitting by creating this low-transmittance trail, we set $\alpha = 1$ at approximately empty regions of space (i.e. $\sigma < 0.05$). However, without the gradient estimator, points with an unfavorable initialization may ``saturate'' $\min(0, \alpha')$ and become stuck at $\alpha = 1$. This can be seen in the pillar in Fig.~\ref{fig:ablation_clip}, which is split into multiple parts when the gradient estimator is not used.
\end{itemize}

\subsection{SSIM by Dataset}
\label{appendix:ssim_full}

Table~\ref{tab:ssim_full} provides a full breakdown of the SSIM of \name\ and each baseline on each view synthesis dataset; \name\ dominates each baseline on all datasets. 95\% (one-sided) confidence intervals for the difference to \name\ are calculated using the procedure described in Sec.~\ref{appendix:metrics}. We also provide the CDF of the SSIM for each dataset as a distributional view (Fig.~\ref{fig:ssim_cdf}).

\subsection{Additional Tomography Examples}
\label{appendix:tomography}

\begin{figure*}
\centering
\begin{subfigure}[b]{0.24\textwidth}
\includegraphics[width=\textwidth]{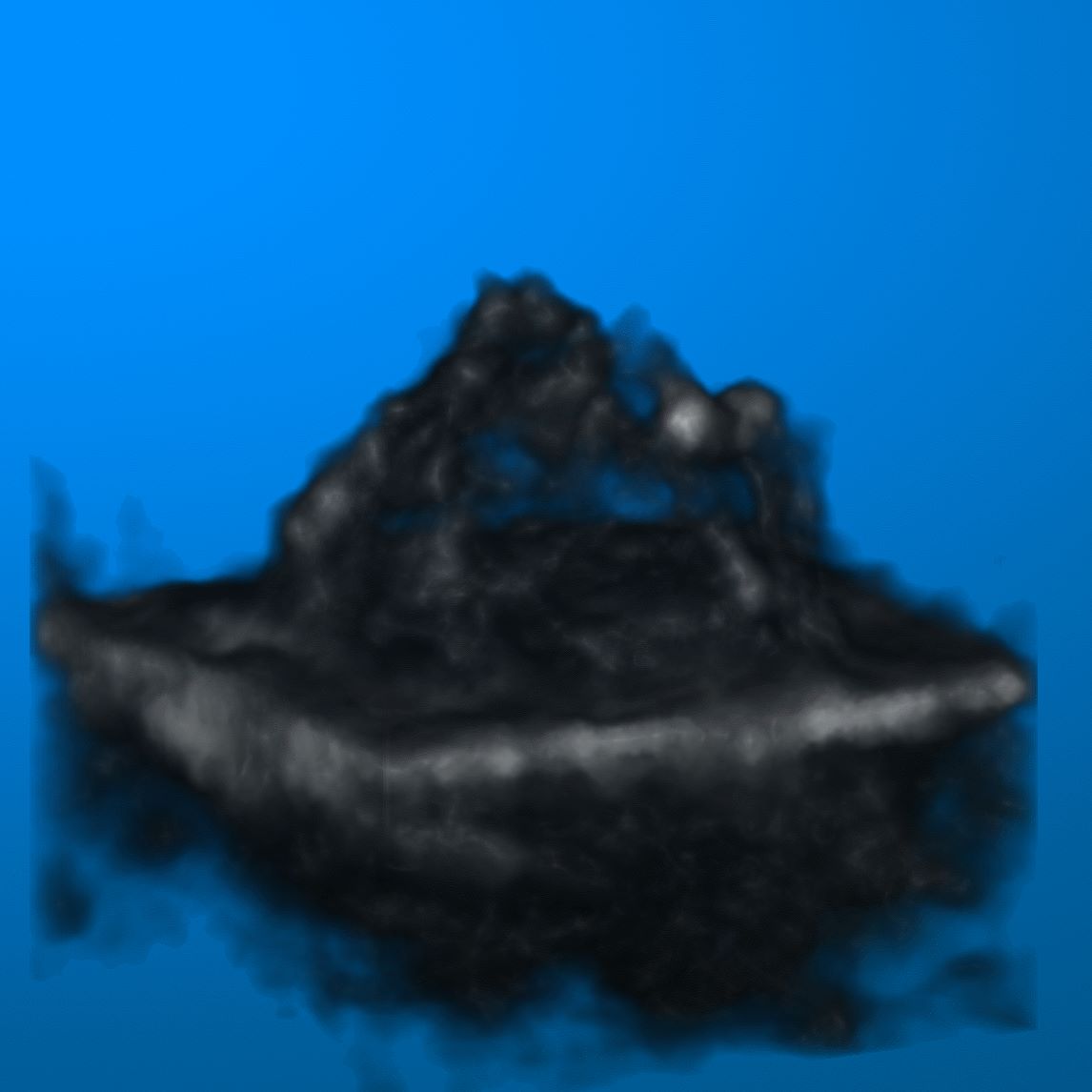}
\caption{Empty - Volume}
\end{subfigure}
\begin{subfigure}[b]{0.24\textwidth}
\includegraphics[width=\textwidth]{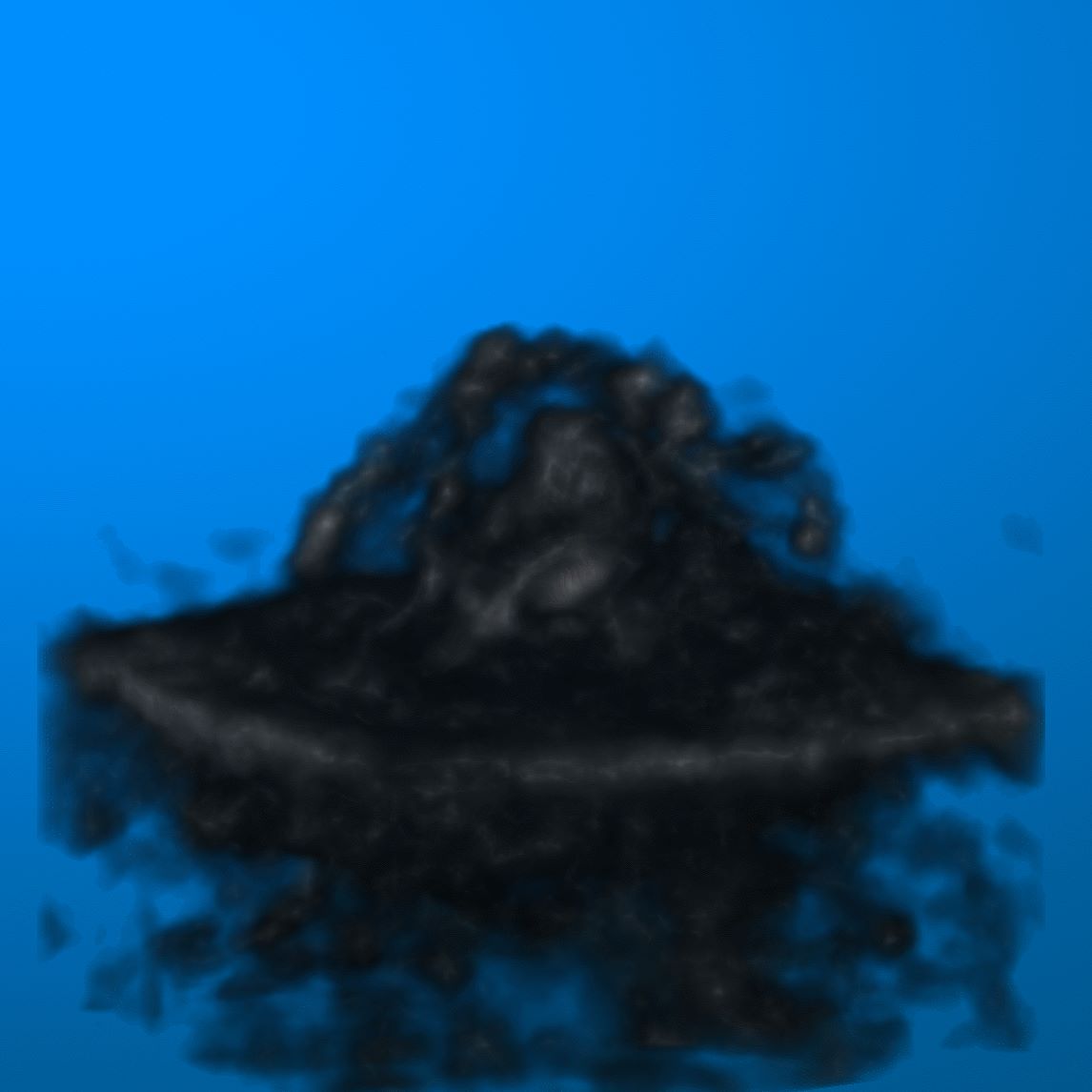}
\caption{Occupied - Volume}
\end{subfigure}
\begin{subfigure}[b]{0.24\textwidth}
\includegraphics[width=\textwidth]{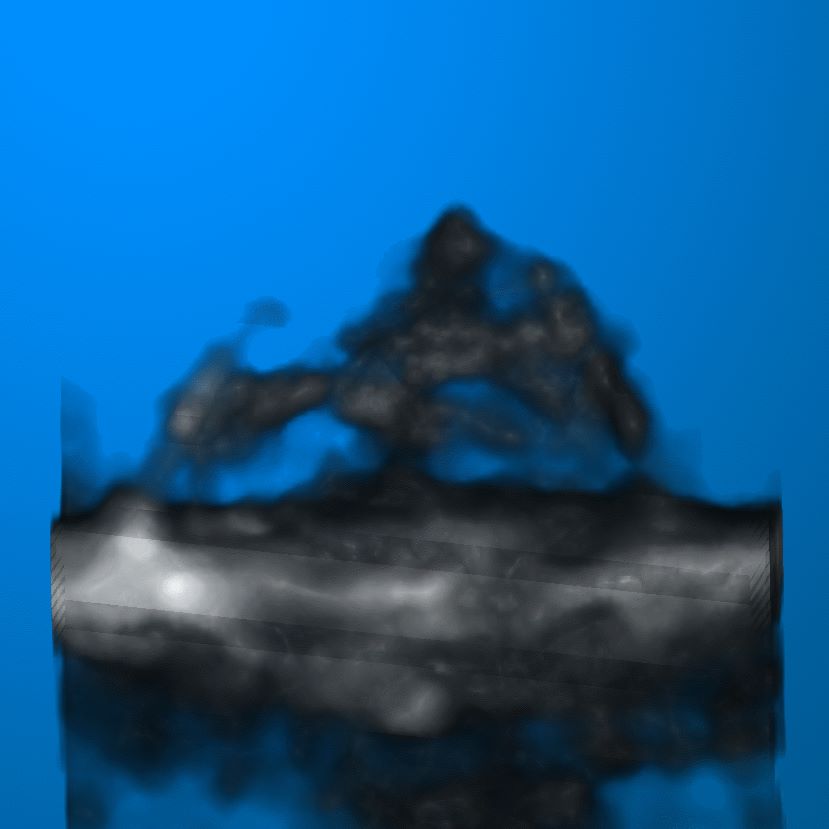}
\caption{Empty - Section}
\end{subfigure}
\begin{subfigure}[b]{0.24\textwidth}
\includegraphics[width=\textwidth]{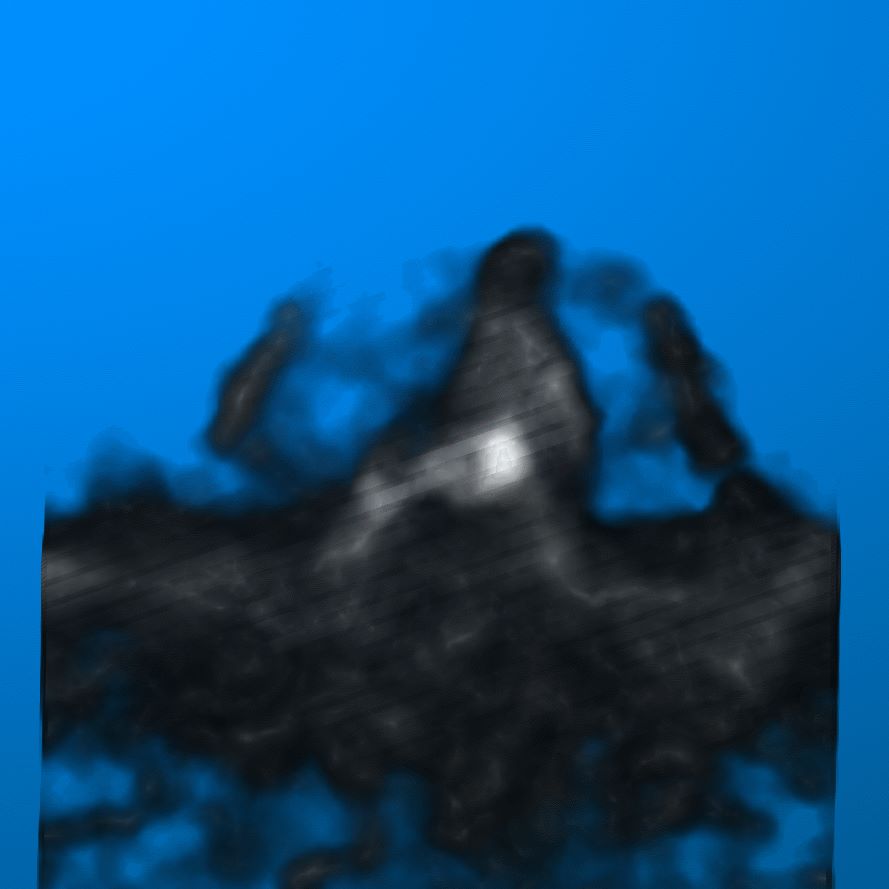}
\caption{Occupied - Section}
\end{subfigure}
\caption{Volume renderings of the tent with and without a person inside, created only using radar sensor readings from outside a tent.}
\label{fig:tent-volume}
\end{figure*}

\begin{figure}
\centering
\includegraphics[height=0.49\columnwidth]{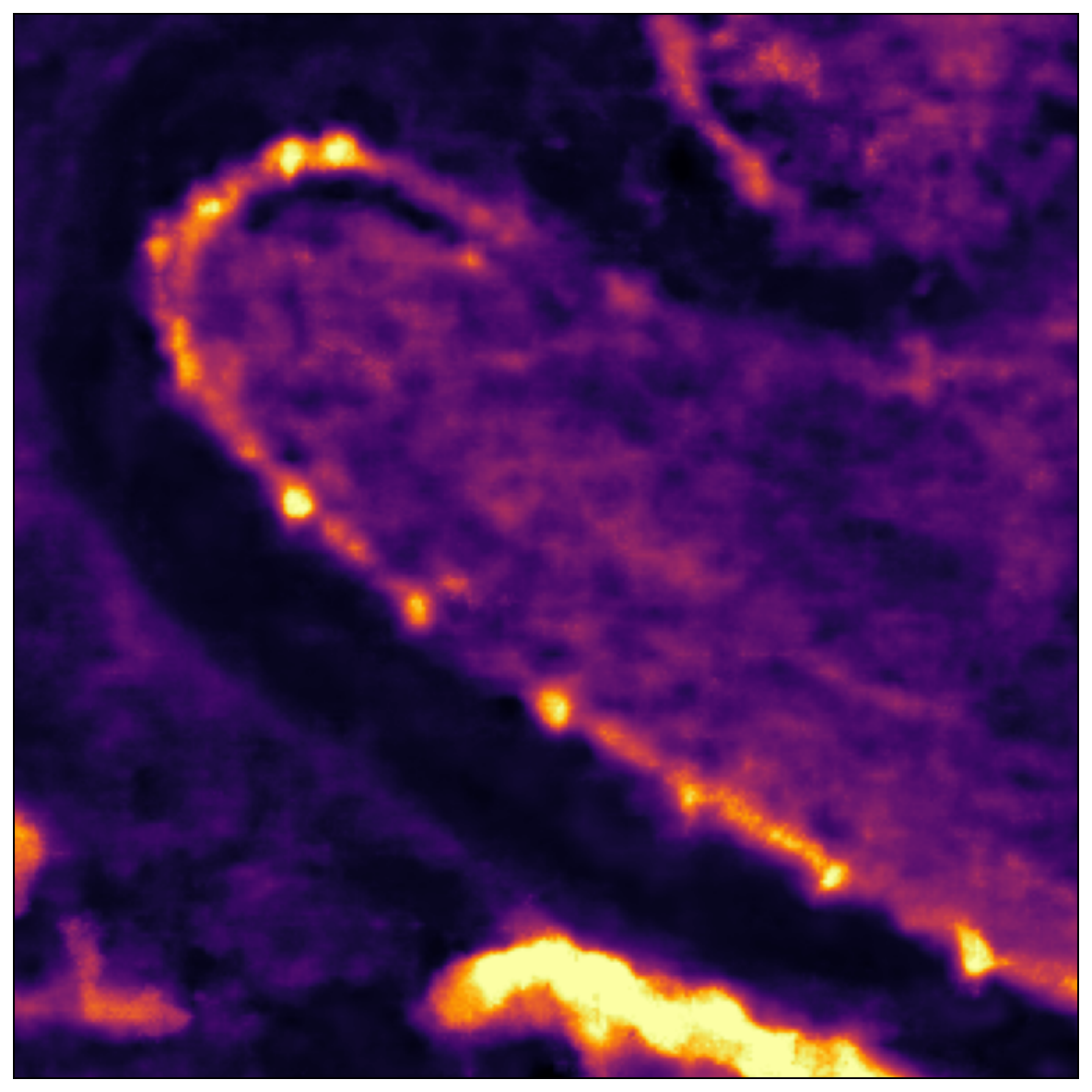}
\includegraphics[height=0.48\columnwidth]{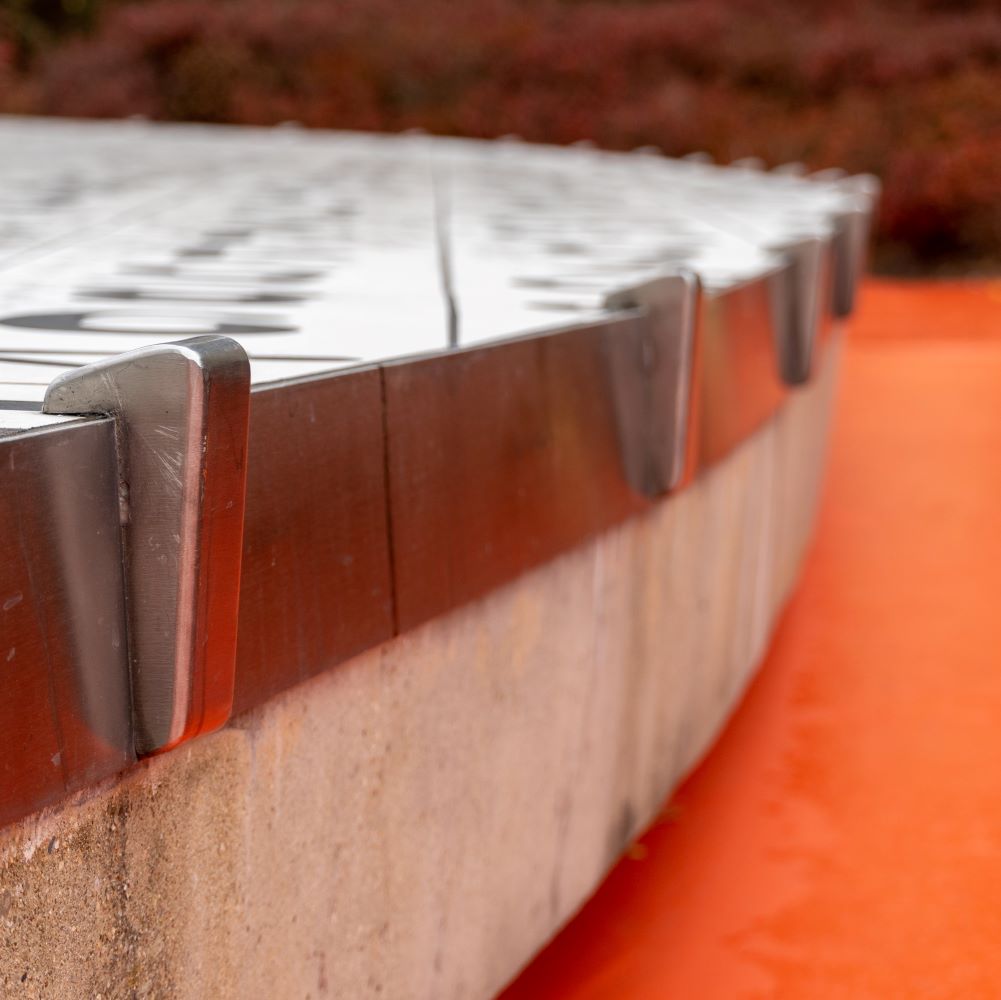}
\vspace{-0.5em}
\caption{Metal ``anti-skate strips'' (the bright dots) are clearly visible in our tomography map along with a metal band on the outside of a raised section (the bright outline of the shape).}
\vspace{-0.5em}
\label{fig:garden-example}
\end{figure}

\paragraph{Garden} Fig.~\ref{fig:garden-example} shows a zoomed view of the \textit{Garden} scene shown in Fig.~\ref{fig:tomography}. Metal features along the edge of the raised section are clearly visible in our tomographic images.

\paragraph{Tent} Fig.~\ref{fig:tent-volume} shows a 3D volume rendering of the \textit{Tent} scene shown in Fig.~\ref{fig:tomography} when occupied and unoccupied. In addition to the tent poles, the tent's occupant is clearly visible; a rough approximation of the occupant's seated pose can also be seen after cutting along a section plane.

\paragraph{Mapping} \name\ can also be used to generate maps of spaces. Fig.~\ref{fig:maps} shows example learned reflectance maps compared to Lidar occupancy grids; \name\ is able to learn a coarse map of the room via its radar reflectance, indicating the potential for future localization and mapping algorithms based on \name.

\end{document}